\documentclass[accepted]{article}

\usepackage{microtype}
\usepackage{graphicx}
\usepackage{booktabs} 

\usepackage{hyperref}

\usepackage{icml2022}

\usepackage{multirow}
\usepackage{amsmath}
\usepackage{amsthm}
\usepackage{amssymb}
\usepackage{bm}
\usepackage{algorithm}
\usepackage{algorithmic}
\usepackage{graphicx,subfig}

\newcommand{\norm}[1]{\left\lVert#1\right\rVert}

\def \R {\mathbb{R}}

\def \vv {\bm{v}}

\newtheorem{assumption}{Assumption}

\newtheorem{lemma}{Lemma}

\newtheorem{theorem}{Theorem}

\newtheorem{remark}{Remark}

\icmltitlerunning{Personalized Federated Learning via Variational Bayesian Inference}

\begin{document}

\twocolumn[
\icmltitle{Personalized Federated Learning via Variational Bayesian Inference}



\icmlsetsymbol{equal}{*}

\begin{icmlauthorlist}
\icmlauthor{Xu Zhang}{equal,AMS}
\icmlauthor{Yinchuan Li}{equal,Huawei}
\icmlauthor{Wenpeng Li}{Huawei}
\icmlauthor{Kaiyang Guo}{Huawei}
\icmlauthor{Yunfeng Shao}{Huawei}
\end{icmlauthorlist}

\icmlaffiliation{AMS}{LSEC, Academy of Mathematics and Systems Science, Chinese Academy of Sciences, Beijing, China}
\icmlaffiliation{Huawei}{Noah’s Ark Lab, Huawei, Beijing, China}

\icmlcorrespondingauthor{Wenpeng Li}{li.wenpeng@huawei.com}

\icmlkeywords{Personalized Federated Learning, Variational Inference, Bayesian
neural network}

\vskip 0.3in
]



\printAffiliationsAndNotice{\icmlEqualContribution} 

\begin{abstract}

Federated learning faces huge challenges from model overfitting due to the lack of data and statistical diversity among clients. To address these challenges, this paper proposes a novel personalized federated learning method via Bayesian variational inference named pFedBayes. To alleviate the overfitting, weight uncertainty is introduced to neural networks for clients and the server. To achieve personalization, each client updates its local distribution parameters by balancing its construction error over private data and its KL divergence with global distribution from the server. Theoretical analysis gives an upper bound of averaged generalization error and illustrates that the convergence rate of the generalization error is minimax optimal up to a logarithmic factor. Experiments show that the proposed method outperforms other advanced personalized methods on personalized models, e.g., pFedBayes respectively outperforms other SOTA algorithms by 1.25\%, 0.42\% and 11.71\% on MNIST, FMNIST and CIFAR-10 under non-i.i.d. limited data.
\end{abstract}

\section{Introduction}

Federated learning (FL) is an increasingly popular topic in deep learning, which can model machine learning for distributed end devices while preserving their privacy \cite{mcmahan2017communication,Li2020FL}. With the increasing emphasis on privacy protection, federated learning has been widely used in finance, medicine, internet of things, internet of vehicles and e-commerce. When data from different clients are assumed to be independent and identically distributed (i.i.d.), federated learning performs well and has a strict convergence guarantee. However, there are two main challenges caused by imperfect data, one is that private data from different clients are usually non-i.i.d. due to the differences in user preferences, locations, and living habits, leading to model performance degradation; the other is that data from clients are usually limited and not enough to train a large neural network with too many parameters, leading to model overfitting. A natural question is can we design a federated learning algorithm to address these two challenges caused by data together?

To overcome this challenge caused by non-i.i.d. data, personalized federated learning (PFL) \cite{li2018federated,t2020personalized,huang2021personalized} is proposed to achieve personalization. Although standard PFL has come a long way in non-i.i.d. data,
model overfitting often occurs when data from the client is limited. Recently, the Bayesian neural network (BNN) is introduced into FL to address the model overfitting by representing all network parameters in the global model with probability distributions \cite{chen2020fedbe,thorgeirsson2021probabilistic}. Unfortunately, these algorithms show performance degradation when data from different clients have statistical diversity. Our goal is to find a way to address the challenges from non-i.i.d. data and limited data simultaneously.

In this paper, we find an efficient way to adapt BNN into PFL and propose a novel PFL model through variational Bayesian inference. Unlike traditional BNNs, we do not assume a prior distribution for each parameter on the end devices; instead, the trained global distribution is used as the prior distribution. It is well known that the assumed prior distribution often does not match the true distribution. Our model can avoid this flaw by using a trained global distribution. Furthermore, the proposed model can quantify the uncertainty of the network output by using Bayesian model averaging, which has practical implications in various robustness-critical applications such as medical diagnosis, autonomous driving, and financial transactions \cite{jospin2020hands}.



\subsection{Main Contributions}

This paper proposes a novel two-level personalized federated learning model named \texttt{pFedBayes} based on variational Bayesian inference. Both local and global neural networks are formulated as BNN, where all network parameters are treated as random variables. The server seeks to minimize the KL divergence between the global distribution and all local distributions. The local model encourages clients to find a local distribution that balances the construction error over its private data and the KL divergence with global distribution. It should be stressed that \texttt{pFedBayes} not only achieves personalization under limited data conditions, but also quantifies the output uncertainty, which can meet the requirements of many practical applications.

Next, we provide the theoretical guarantee for the proposed model \texttt{pFedBayes}. We give the upper bound on the averaged generalization error across all clients, which shows that the upper bound consists of estimation error and approximation error and the estimation error is on the order of $1/n$. Furthermore, by choosing a suitable number of network parameters, we prove that the averaged generalization error achieves {\em minimax optimal} convergence rate up to a logarithmic term.

Finally, we propose a computationally efficient algorithm by using the one-order stochastic gradient algorithm for clients and the server alternatively and compare the performance of \texttt{pFedBayes} with other state-of-the-art (SOTA) algorithms under statistical diversity conditions. Various experimental results present that the proposed algorithm has a better performance than other SOTA personalization algorithms, especially in the case of limited data. In particular, \texttt{pFedBayes} has the highest accuracy on  personalized models on MNIST, FMNIST and CIFAR-10 datasets with small, medium, and large data volumes, and  \texttt{pFedBayes} has competitive performance on the global models when the amount of data is small and medium.

\subsection{Related Works}
Personalized Bayesian federated learning is closely related to the following topics:

\textbf{Federated learning.}  Google group proposed the first federated learning algorithm named FedAvg (Federated Averaging) to protect the privacy of clients in distributed learning \cite{mcmahan2017communication}. 
Many variants of FedAvg were proposed to meet different demands from various scenarios. To reduce the round of communication, communication-efficient algorithms were considered in federated learning such as approximate Newton's algorithm \cite{li2019feddane}, primal-dual algorithm \cite{zhang2020fedpd} and one-shot averaging algorithm \cite{guha2019one}. To reduce the data size of storage and communication, sparsity \cite{sattler2019robust,rothchild2020fetchsgd} and quantization \cite{dai2019hyper,reisizadeh2020fedpaq,zong2021communication} were studied in federated learning. These algorithms focus on learning a global model for all clients, which presents poor performance when private data from different clients are non-i.i.d.

\textbf{Personalized federated learning.} To address the challenge from heterogeneous datasets, researchers proposed a number of PFL methods including local customization methods \cite{li2018federated,arivazhagan2019federated, hanzely2020federated,t2020personalized,huang2021personalized,li2021personalized,liu2022sparse}, 
multi-task learning based methods \cite{smith2017federated,sattler2021clustered}, meta-learning based methods \cite{chen2018federated,fallah2020personalized} and others \cite{li2022federated,li2022mining}. Local customization methods customize a personalized model for each client. For example, \citet{li2018federated} proposed a variant of FedAvg named Fedprox by adding a proximal term in the subproblems; 
\citet{huang2021personalized} proposed FedAMP and HeurFedAMP by designing an attentive message passing mechanism to facilitate more collaborations of similar clients. Two-level modeling is a special kind of customization method, which is composed of the sever-level subproblem and client-level subproblems. In particular, pFedMe \cite{t2020personalized} penalized the $\ell_2$ norm of the difference between the local parameter and global parameter in the client-level subproblem, which has a similar two-level optimization problem with pFedBayes but has a poor performance for small datasets.
For the multi-task learning based method, \citet{smith2017federated} first introduced multi-task learning into federated learning and proposed a robust optimization method. Then \citet{sattler2021clustered} proposed a clustered federated learning to realize personalization by using the geometric properties of the loss function. Regrading the meta-learning based methods, \citet{fallah2020personalized} considered a variant of FedAvg named per-FedAvg by jointly obtaining an initial model and then applying it to each client. Although these advanced algorithms improve the performance on non-i.i.d. data, they suffer from overfitting when the amount of data is limited.

\begin{figure*}[!th]
\hspace{0.35cm}
    \begin{minipage}{0.50\linewidth}
      \centering
      \centerline{\includegraphics[width=1.1\linewidth]{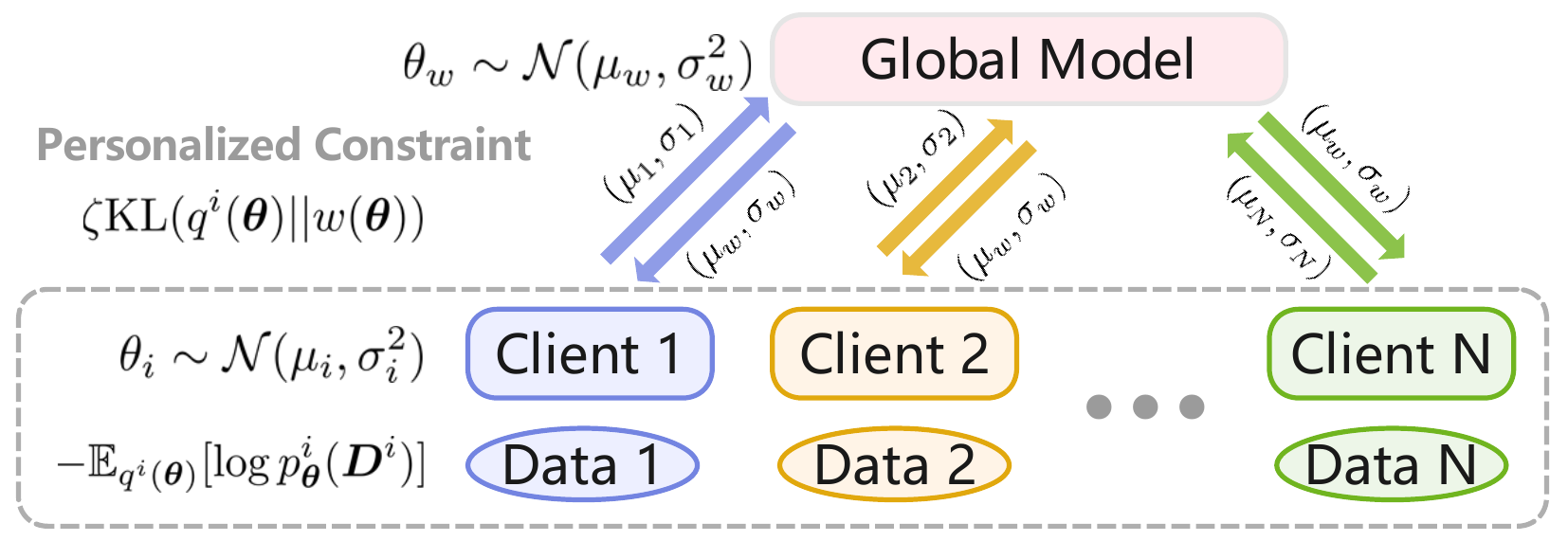}}
    \end{minipage}
    \hfill
   \begin{minipage}{0.50\linewidth}
      \centering
        \centerline{\includegraphics[width=0.9\linewidth]{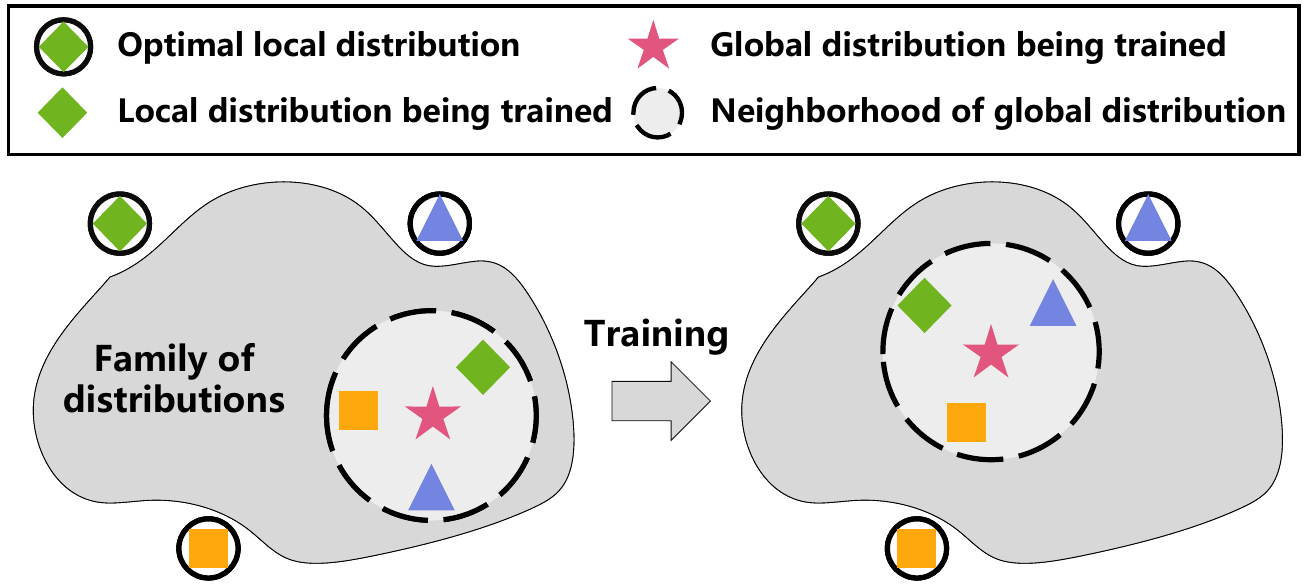}}
    \end{minipage}
\caption{Personalized Bayesian federated learning model under Gaussian assumptions. \textbf{Left:} System diagram. Each client uploads its updated distribution to the server and then downloads the aggregated global distribution from the server. \textbf{Right:} Distribution Training. The subfigure shows the evolution of training the local and global distributions in probability space.}
    \label{fig:fig1}
\end{figure*}

\textbf{Bayesian neural network and Bayesian federated learning.} To overcome the overfitting caused by limited data, the Bayesian neural network \cite{mackay1992practical, neal2012bayesian, blundell2015weight} was proposed by imposing a prior distribution on each parameter (weights and biases). The authors in \cite{pati2018statistical,cherief2018consistency,maddox2019simple, alquier2020concentration,bai2020efficient, cherief2020convergence} studied the generalization error bounds and gave the concentration of variational approximation for different statistical models. 
To alleviate the overfitting in FL, a Bayesian ensemble was introduced into FL.  \citet{yurochkin2019bayesian} proposed a Bayesian nonparametric federated learning (BNFed) framework by neural parameter matching based on the Beta-Bernoulli process. \citet{chen2020fedbe} proposed FedBE by applying the Bayesian model to the global model and treating local distributions as Gaussian or Dirichlet distributions. Based on FedAvg, \citet{thorgeirsson2021probabilistic} introduced Gaussian distribution to each parameter and aggregated the local models by calculating the mean and variance of uploaded parameters. However, the above models have poor performance when data are non-i.i.d. \citet{al2021federated} proposed an approximate posterior inference algorithm named FedPA by inferring global posterior via the average of local posteriors, but FedPA is inapplicable for PFL. \citet{achituve2021personalized} proposed a personalized model based on Gaussian processes named pFedGP, which learns a shared deep kernel function for all clients and a personalized Gaussian process classifier using a personal dataset for each client. Since all the clients use a common kernel function, the performance might degrade when data varies widely. Different from pFedGP, each client in our model owns its personalized BNN, which is more flexible and robust in practical applications. \citet{liu2021bayesian} proposed an FL model named FOLA to achieve personalization, where Gaussian approximate posterior distribution of the server parameter is considered as the client's prior. But FOLA is based on Maximum a posteriori estimation while the pFedBayes is based on a Bayesian two-level optimization. Furthermore, FOLA lacks theoretical analysis, but pFedBayes has theoretical guarantees.




\section{Personalized Bayesian Federated Learning Model with Gaussian Distribution}

In this section, we present the personalized Bayesian federated learning model with Gaussian distribution. In our model, each client owns its BNN, where each parameter (weight or bias) denotes a Gaussian distribution. Compared with the classical neural network, we only use twice the number of parameters to construct a BNN, but includes an infinite number of classical neural network. As shown in Fig. \ref{fig:fig1}(a), at each iteration, the clients upload their distributions (i.e., mean and variance) to the server and then download the aggregated distribution from the server. In Fig. \ref{fig:fig1}(b), we show the evolution of the distribution training, where the global model (GM) serves as the prior of the personalized models (PMs). The figure presents that the trained distribution for each PM lies between the GM and PM without communication, which balances the GM and extreme PM without communication.

Next, we show the proposed model explicitly. Considering a distributed system that contains one server and $N$ clients. For ease of analysis, we assume the variance of noise for all clients is the same and the number of data is the same. Let the $i$-th client satisfy the model
\begin{equation}
    \bm{y}_j^i=f^i(\bm{x}_j^i)+\varepsilon_j^i,\, j=1,\ldots,n,\, \varepsilon_j^i \sim \mathcal{N}(0,\sigma_\varepsilon^2),
\end{equation}
where $\bm{x}_j^i\in \R^{s_0}$, $\bm{y}_j^i \in \R^{s_{L+1}}$ for $j=1,\ldots,n$, $i=1,\ldots,N$, $f^i(\cdot): \R^{s_0} \to \R^{s_{L+1}}$ denotes a nonlinear function, $n$ denotes the sample size and $\sigma_\varepsilon$ denotes the variance of noise.  Define the dataset for $i$-th client as  $\bm{D}^i=(\bm{D}_1^i,\ldots,\bm{D}_n^i)$, where $\bm{D}_j^i=(\bm{x}_j^i,\bm{y}_j^i)$. Let $P^i$ denote the probability measure of data for the $i$-th client and $p^i$ be its corresponding probability density function.

Suppose each client has the same neural network architecture, i.e., a fully-connected Deep Neural Network (DNN),  but has different parameters due to the fact that their data are non-i.i.d. In particular, the neural network has $L$ hidden layers, where the $j$-th hidden layer has $s_j$ neurons and activation functions $\sigma(\cdot), j=1,\ldots,L$. The output of the DNN model is represented as $f_{\bm \theta}(\bm{x})$, where $\bm{\theta} \in \R^T$ denotes the vector that contains all parameters and $T$ is the length of $\bm{\theta}$. Define $\bm{s}=\{s_1,\ldots,s_L\}$. Assume that all parameters are bounded, i.e., there exixts some constant $B>0$ such that $\norm{\bm{\theta}}_\infty \le B$. Then we denote the DNN model for the $i$-th client as $f_{\bm{\theta}}^i$.


Our goal is to design an FL model that achieves personalization and alleviates overfitting simultaneously. 
Before giving a personalized Bayesian FL model, we first review the standard BNN model via variational inference \cite{VIJordan1999,Blei2017ReviewVB}. The optimization problem aims to find the closest distribution to the posterior distribution in the variational family of distributions $\mathcal{Q}$ 
\begin{equation}
 \min_{q( \bm{\theta}) \in \mathcal{Q}} \mbox{KL}(q( \bm{\theta})||\pi (\bm{\theta}|\bm{D})),
\end{equation}
where $\pi (\bm{\theta}|\bm{D})$ denotes the posterior distribution and $\bm{D}$ denotes the collected data. Using Bayes theorem gives $\pi (\bm{\theta}|\bm{D}) \propto \pi (\bm{\theta}) p_{\bm{\theta}} (\bm{D})$, where $\pi (\bm{\theta})$ denotes the prior distribution and $p_{\bm{\theta}} (\bm{D})$ denotes the likelihood.
So the above optimization problem is equivalent to the following problem
\begin{equation}
 \min_{q( \bm{\theta}) \in \mathcal{Q}} - \mathbb{E}_{q(\bm{\theta})}[\log p_{\bm{\theta}}(\bm{D})] +\mbox{KL}(q(\bm{\theta})||\pi (\bm{\theta})),
\end{equation}
where the first term can be regarded as the reconstruction error on the dataset $\bm{D}$ and the second term is a regularization with the prior distribution.

Based on the standard BNN model, we propose a novel personalized federated learning model via variational Bayesian inference named \texttt{pFedBayes}, which is formulated as the following two-level optimization problem
\begin{align}
\label{Master-1}
    \text {Server:}&~ \min _{w( \bm{\theta}) \in \mathcal{Q}_w}\left\{F(w)\triangleq \frac{1}{N} \sum_{i=1}^{N} F_{i}(w)\right\},
\end{align}
\begin{align}
\label{Client-1}
   \text {Clients:}~ F_i(w)\triangleq \min _{q^i(\bm{\theta}) \in \mathcal{Q}^i}\Big\{&- \mathbb{E}_{q^i(\bm{\theta})}[\log p^i_{\bm{\theta}}(\bm{D}^i)] 
   \nonumber\\
   & + \zeta\, \mbox{KL}(q^i(\bm{\theta})||w(\bm{\theta})) \Big\}.
\end{align}
where $w(\bm{\theta})$ and $q^i(\bm{\theta})$ denote the global distribution and the local distribution for the $i$-th client to be optimized, respectively, $\mathcal{Q}_w$ and $\mathcal{Q}^i$ denote the family of distributions for global parameter and the $i$-th client, respectively,  and $p^i_{\bm{\theta}}(\bm{D}^i)$ denotes the likelihood for the $i$-th client. By minimizing the sum of KL divergence, the global model can find the closest distribution in $Q_w$ to the clients' distribution. Note that in the regularization term for clients, we replace prior distribution with global distribution. The reason is that since we cannot characterize the prior distribution in practice, it's difficult to assume a good prior distribution that is compatible with the collected data. Instead, by replacing the prior distribution with trained global distribution, we find a relatively good distribution without making assumptions about the prior distribution. Motivated by the modification in $\beta$-VAE \cite{higgins2017beta}, there is an extra parameter $\zeta \ge 1$ in the subproblem that balances the degree of personalization and global aggregation.

In this paper, we assume that the parameters of the neural network follow Gaussian distribution, which is commonly used in the literature \cite{Blundell2015,chen2020fedbe,thorgeirsson2021probabilistic}. Besides, it's common to assume that the distribution satisfies mean-field decomposition, that is, the joint distribution equals to the product of each parameter's distribution. In particular, assume that the family of distributions of the $i$-th client $\mathcal{Q}^i$ satisfies 
\begin{equation}\label{ptheta}
\theta_{i,m}  \sim  \mathcal{N}(\mu_{i,m}, \sigma^2_{i,m})  , ~m=1,\ldots, T, 
\end{equation}
where $\mu_{i,m}$ denotes the Gaussian mean and  $\sigma^2_{i,m}$ denotes the Gaussian variance for $m$-th parameter of the $i$-th client. Let the family of distributions of the server $\mathcal{Q}_w$ follow 
\begin{equation}\label{wstar}
\theta_{w,m} \sim \mathcal{N}(\mu_{w,m}, \sigma^2_{w,m}) ,~m=1,\ldots, T,
\end{equation}
where $\mu_{w,m}$ denotes the Gaussian mean and  $\sigma^2_{w,m}$ denotes the Gaussian variance for $m$-th parameter of the server. 

Armed with the above distributions, we can obtain the KL divergences of the two Gaussian distributions $q^i(\bm{\theta})$ and $w(\bm{\theta})$ as follows
\begin{align} \label{kl_q_w}
    &~\mbox{KL}(q^i(\bm{\theta})||w(\bm{\theta})) \nonumber\\
    =&~\mbox{KL}\left(\prod^{T}_{m=1}\mathcal{N}(\mu_{i,m}, \sigma^2_{i,m}) \Bigl|\Bigr| \prod^T_{m=1}\mathcal{N}(\mu_{w,m}, \sigma^2_{w,m})\right) \nonumber\\
    =&~\sum_{m=1}^T  \mbox{KL}(\mathcal{N}(\mu_{i,m}, \sigma^2_{i,m})||\mathcal{N}(\mu_{w,m}, \sigma^2_{w,m}))  \\
    =&~ \frac{1}{2} \sum_{m=1}^{T} 
    \left[ \log \left( \frac{\sigma_{w,m}^2}{\sigma_{i,m}^2} \right) + \frac{\sigma_{i,m}^2+ (\mu_{i,m}-\mu_{w,m})^2}{\sigma_{w,m}^2} -1
    \right].\nonumber
\end{align}


\section{Theoretical Analysis}

In this section, we will provide theoretical analysis for averaged generalization error of the proposed model and show the minimax optimality of the convergence rate of the generalization error. The proofs of the main results are put in the Appendices. 

Before that, we give some necessary assumptions. For simplicity, we analyse the equal-width Bayesian neural network as \citet{polson2018posterior} and \citet{bai2020efficient}. And we study a general activation function that satisfies $1$-Lipschitz continuous.

\begin{assumption} \label{assump: equal-width}
The widths of neural network are equal-width, i.e.,  $s_i=M.$
\end{assumption}

\begin{assumption} \label{assump: non-linear}
The  activation function $\sigma(\cdot)$ is $1$-Lipschitz continuous.
\end{assumption}

\begin{assumption} \label{assump: sigma_bound}
The parameters $s_0,n,M,L$  are large enough such that 
\begin{equation}
    \sigma^2_{n}=  \frac{T}{8n} A \le B^2,
\end{equation}
where $H=BM$ and 
\begin{align} \label{eq: A}
&A=\log^{-1}(3s_0M)\cdot(2H)^{-2(L+1)} \nonumber\\ &\left[\left(s_0 + 1 + \frac{1}{H-1}\right)^2 + \frac{1}{(2H)^2-1}
	+\frac{2}{(2H-1)^2}\right]^{-1}.
\end{align}
\end{assumption}

\begin{remark} This sequence ${\sigma_n^2}$ is constructed to prove Lemma \ref{lm: upper bound 2}. Since the neural parameters are bounded by $B$, their variance should be upper bounded by $B^2$. The above assumption is easy to be satisfied due to the fact that $A/8n$ can be arbitrarily small. 
\end{remark}

Next, we give some useful definitions. Define the Hellinger distance as follows
$$d^2(P_{\bm{\theta}}^i, P^i) = \mathbb{E}_{X^i}\Bigl( 1 - \exp \{-[f_{\bm{\theta}}^i(X^i) - f^i(X^i)]^2/(8\sigma^2_{\epsilon}) \}\Bigr).$$

Define the following terms
\begin{align}
r_n&= ((L+1)T/n)\log M + (T/n)\log(s_0\sqrt{n/T}),\\
\xi^i_n&= \inf_{\bm{\theta} \in {\Theta}(L, \bm{s}),\|\bm{\theta}\|_\infty\leq B}||f_{\bm{\theta}}^i-f^i||^2_\infty,\\
\varepsilon_n &= n^{-\frac{1}{2}}\sqrt{(L+1)T\log M + T\log(s_0\sqrt{n/T})}\log^\delta(n),
\end{align}
where $\delta>1$.

Since the incorporation of a constant that is unrelated to $\bm{\theta}$ doesn't affect the optimization problem, we rewrite the subproblem for clients as follows
\begin{align}
\label{Client-2}
   \text {Clients:}~ F_i(w)\triangleq \min _{q^i(\bm{\theta}) \in \mathcal{Q}^i} & \Big\{  \int_{\Theta} l_n(P^{i},P^{i}_{\bm{\theta}}) q^i(\bm{\theta}) d\bm{\theta} \nonumber\\
   &+ \zeta\,\mbox{KL}(q^i(\bm{\theta})||w(\bm{\theta})) \Big\},
\end{align}
where $l_n(P^{i},P^{i}_{\bm{\theta}})$ is the log-likelihood ratio of $P^{i}$ and $P^{i}_{\bm{\theta}}$
\begin{equation}
    l_n(P^{i},P^{i}_{\bm{\theta}})=\log \frac{p^i(\bm{D}^i)}{p^{i}_{\bm \theta}(\bm{D}^i)}.
\end{equation}

Let $w^\star(\bm{\theta})$ be the optimal variational solution of the problem and $\hat{q}^i(\bm{\theta})$ be its corresponding variational solution for the $i$-th client's subproblem
\begin{align} \label{prb: subproblem_optimal}
    \hat{q}^i(\bm{\theta})
    =\arg \inf_{q^i(\bm{\theta}) \in \mathcal{Q}^i} \Big\{ &\int_{\Theta} l_n(P^{i},P^{i}_{\bm{\theta}}) q^i(\bm{\theta}) d\bm{\theta} \nonumber\\
    &+ \zeta\, \mbox{KL}(q^i(\bm{\theta})||w^\star(\bm{\theta}))\Big\}.
\end{align}
Our purpose is to give an upper bound for the term
\begin{align} \label{dist}
  \frac{1}{N} \sum_{i=1}^{N} \int_{\Theta}   d^2(P_{\bm{\theta}}^i, P^{i}) \hat{q}^i(\bm{\theta}) d\bm{\theta}. 
\end{align}
with parameters $r_n$, $\xi^i_n$ and $\varepsilon_n$.




We first bound Eq. \eqref{dist} by the average of objective functions of all subproblems as follows.

\begin{lemma} \label{lm: upper bound 1}
Suppose that the assumptions 1 and 2 are true, then with dominating probability, the following inequality holds
\begin{align} \label{neq: local_upperbound_sum}
    \frac{1}{N} & \sum_{i=1}^{N}  \int_{\Theta}  d^2(P_{\bm{\theta}}^i, P^i)  \hat{q}^i(\bm{\theta}) d\bm{\theta} \le \nonumber\\
    &\frac{1}{n} \Bigg\{\frac{1}{N} \sum_{i=1}^{N} \Bigg[ \frac{1}{\zeta} \int_{\Theta} l_n(P^{i}, P^{i}_{\bm{\theta}})\hat{q}^i(\bm{\theta}) d\bm{\theta}  \nonumber\\
    & \qquad \qquad \quad + \mbox{\rm KL}(\hat{q}^i(\bm{\theta})||w^\star(\bm{\theta})) \Bigg] \Bigg\} + C\varepsilon^{2}_{n}, 
\end{align}
where $\zeta \ge 1$ is a tradeoff parameter and $C>0$ is an absolute constant.
\end{lemma}

Then we present the upper bound of the average of objective functions for $N$ subproblems.
\begin{lemma} \label{lm: upper bound 2}
Suppose that the assumptions are true, then with dominating probability, the following inequality holds
\begin{multline}
    \frac{1}{N} \sum_{i=1}^{N}  \left[  \int_{\Theta} l_n(P^{i},P^{i}_{\bm{\theta}}) \hat{q}^i(\bm{\theta}) d\bm{\theta} +  \zeta\, \mbox{\rm KL}(\hat{q}^i(\bm{\theta})||w^\star(\bm{\theta}))\right] 
    \\ \le n\left(  C' \zeta r_n +\frac{C''}{N} \sum_{i=1}^{N}\xi^i_n \right),
 \end{multline}
 where $\zeta \ge 1$ is a tradeoff parameter and $C', C''$ are any diverging sequences.
\end{lemma}

Combining Lemmas \ref{lm: upper bound 1} and \ref{lm: upper bound 2} yields the main theorem.

\begin{theorem} \label{Thm_main}
Suppose that the assumptions are true, then the following upper bound holds with dominating probability
\begin{multline}
\label{neq: local_upperbound}
  \frac{1}{N} \sum_{i=1}^{N} \int_{\Theta}  d^2(P_{\bm{\theta}}^i, P^i) \hat{q}^i(\bm{\theta}) d\bm{\theta}
    \\ \le  C \varepsilon^{2}_{n}
  + C'  r_n +\frac{C''}{N\zeta} \sum_{i=1}^{N}\xi^i_n,
\end{multline} 
where $\zeta \ge 1$ is a tradeoff parameter, $C>0$ is an absolute constants and $C', C''$ are any diverging sequences. 
\end{theorem}

The upper bound can be divided into two parts: the first and second terms belong to the estimation error while the third term is the approximation error. Note that the estimation error decreases with the increase of sample size $n$. For the approximation error, it is only related to the total number of parameters $T$ (or the width and depth) of the neural network. Besides, with the increase of $T$, the estimation error increases but the approximation error decreases. Therefore, a suitable parameter $T$ should be chosen as a function of sample size $n$ to balance the upper bound.  

Next, we present the choice of $T$ for a typical class of functions. Assume that $\{f^i\}$ are $\beta$-Hölder-smooth functions and the intrinsic dimension of data is $d$. According to Corollary 6 in \cite{nakada2020}, the approximation error is bounded as follows
\begin{equation}
    \xi^i_n \le C_0 T^{-2\beta/d}, i=1,\ldots,N,
\end{equation}
where $C_0>0$ is a constant related to $s_0$, $\beta$ and $d$.
By using Theorem \ref{Thm_main} and choosing $T=C_1 n^{d/(2\beta+d)}$, we can give the convergence rate of the upper bound
\begin{align} \label{eq: convegencerate}
     \frac{1}{N} \sum_{i=1}^{N} \int_{\Theta}  d^2(P_{\bm{\theta}}^i, P^i) \hat{q}^i(\bm{\theta}) d\bm{\theta} \le C_2 n^{-\frac{2\beta}{2\beta+d}} \log^{2\delta'} (n),
\end{align}
where $\delta'>\delta>1$,  and $C_1, C_2>0$ are constants related to $s_0$, $\beta$, $d$, $L$, $M$, $\zeta$ and $n$. 

Finally, we show the minimax optimality of the generalization error rate. Similar to Theorem 1.1 from \cite{bai2020efficient}, we can represent the convergence result under $L^2$ norm. By using the monotonically decreasing property of $[1-\exp(-x^2)]/{x^2}$ when $x>0$, for bounded  functions $\norm{f^i}_\infty \le F$ and $\norm{f_{\bm{\theta}}^i}_\infty \le F$, $i=1,\ldots,N$, we have 
\begin{align}
    \frac{d^2(P_{\bm{\theta}}^i, P^i)}{\norm{f_{\bm{\theta}}^i(X^i) - f^i(X^i)}_{L^2}^2} \ge \frac{1-\exp(-\frac{4F^2}{8\sigma_\epsilon^2})}{4F^2} \triangleq C_F.
\end{align}
Then we can give an upper bound under $L^2$ norm like \eqref{eq: convegencerate}
\begin{multline}
\label{neq: local_upperbound_L_2}
  \frac{C_F}{N} \sum_{i=1}^{N} \int_{\Theta} \norm{f_{\bm{\theta}}^i(X^i) - f^i(X^i)}_{L^2}^2 \hat{q}^i(\bm{\theta}) d\bm{\theta} \\
  \le \frac{1}{N} \sum_{i=1}^{N} \int_{\Theta}  d^2(P_{\bm{\theta}}^i, P^i) \hat{q}^i(\bm{\theta}) d\bm{\theta} 
    \\ \le  C_2 n^{-\frac{2\beta}{2\beta+d}} \log^{2\delta'} (n).
\end{multline} 

According to the minimax lower bound under $L^2$ norm in Theorem 8 of \cite{nakada2020}, we obtain
\begin{multline}
\label{neq: local_lowerbound_L}
 \inf_{\{\norm{f_{\bm{\theta}}^i}\le F\}_{i=1}^{N}} \sup_{\{\norm{f^i}_\infty\le F\}_{i=1}^N} \frac{C_F}{N} \sum_{i=1}^{N} \\ \int_{\Theta} \norm{f_{\bm{\theta}}^i(X^i) - f^i(X^i)}_{L^2}^2  \hat{q}^i(\bm{\theta}) d\bm{\theta}  \ge C_3 n^{-\frac{2\beta}{2\beta+d}},
\end{multline} 
where $C_3>0$ is a constant.

Combining Eqs. \eqref{neq: local_upperbound_L_2} and \eqref{neq: local_lowerbound_L}, we conclude that the convergence rate of the generalization error of \texttt{pFedBayes} is minimax optimal up to a logarithmic term for bounded functions $\{f^i\}_{i=1}^{N}$ and $\{f^i_{\bm{\theta}}\}_{i=1}^{N}$.

\begin{remark} According to Theorem \ref{Thm_main}, we notice that with the increase of $\zeta$, the approximation error in the upper bound decreases, so does the upper bound. But from the formulation of the proposed optimization problem, we know the increase of $\zeta$ will decrease the degree of personalization. So a suitable choice of $\zeta$ is necessary to balance the degree of personalization and the value of the global upper bound.
\end{remark}



\begin{algorithm}[!t] \small

\begin{minipage}{\linewidth}
	\caption{{\texttt{pFedBayes}}: Personalized Federated Learning via Bayesian Inference Algorithm}
	\begin{tabular}{l} \label{Algorithm_PerFedBayes}
{\bf{Cloud server executes:}} \\
	\hspace{0.4cm}\bf{Input} $T, R, S, \lambda, \eta, \beta, b, \bm v^{0}=(\bm \mu^0,\bm\sigma^0)$ \\
	\hspace{0.4cm}\bf{for} $t=0,1,...,T-1$ \bf{do} \\
      	\hspace{0.8cm}\bf{for} $i=1,2,...,N$ \bf{in parallel do} \\
			\hspace{1.5cm}$\bm v_i^{t+1} \leftarrow \text{ClientUpdate}(i,\bm v^t)$  \\
		\hspace{0.8cm}$\mathbb{S}^{t} \leftarrow \text{Random subset of clients with size~} S$ \\
		\hspace{0.8cm}$\bm v^{t+1}=(1-\beta) \bm v^{t}+\frac{\beta}{S} \sum_{i \in S^t} {\bm v_i^{t+1}}$
	\\
	$\textbf{ClientUpdate}(i,\bm v^{t})\textbf{:}$ \\
		\hspace{0.4cm}$ \bm v_{w,0}^{t} = \bm v^{t}$  \\
		    \hspace{0.4cm}{\bf{for}} $r=0,1,...,R-1$ \bf{do}  \\
		        \hspace{0.8cm}$\bm{D}^{i}_\Lambda$ $\leftarrow$ sample a minibatch $\Lambda$ with size $b$ from $\bm{D}^i$ \\
		        \hspace{0.8cm} $\bm{g}_{i,r}$ $\leftarrow$ Randomly draw $K$ samples from $\mathcal{N}(0,1)$\\
		        \hspace{0.8cm} $\Omega^i(\bm v_{r}^t)$ $\leftarrow$  Use (\ref{eq: generation}) and  (\ref{Client-FedBayes-2}) with $\bm{g}_{i,r}$, $\bm{D}^{i}_\Lambda$ and $\bm v_{r}^t$\\
		        \hspace{0.8cm} $\nabla_{\bm{v}} \Omega^i (\bm v_{r}^t)$ $\leftarrow$  Back propagation w.r.t $\bm v_{r}^t$ \\
		         \hspace{0.8cm} $\bm v_{r}^t$ $\leftarrow$ Update with $\nabla_{\bm{v}} \Omega^i(\bm v_{r}^t)$ using GD algorithms \\
		         \hspace{0.8cm} $\Omega^i_w(\bm v_{w,r}^t)$ $\leftarrow$  Forward propagation w.r.t $\bm v$\\
		        \hspace{0.8cm} $\nabla \Omega^i_w(\bm v_{w,r}^t)$ $\leftarrow$  Back propagation w.r.t $\bm v$\\
		         \hspace{0.8cm} Update $\bm v_{w,r+1}^t$ with $\nabla \Omega^i_w(\bm v_{w,r}^t)$ using GD algorithms \\
		\hspace{0.4cm} return $\bm{v}_{w,R}^{t}$ to the cloud server
	\end{tabular}
\end{minipage}
\end{algorithm}

\section{Algorithm}

In this section, we will show how to implement the model in (\ref{Master-1}) and (\ref{Client-1}) via first-order stochastic gradient decent (SGD) algorithms. Instead of using $\bm{\theta}$ in the optimization problem, we use two new vectors $\bm{\mu}$ and $\bm{\rho}$.
Here, $\mu_m$ denotes the mean and $\sigma_m=\log(1+\exp(\rho_m))$ denotes the standard deviation of random variable ${\theta}_m$. The introduction of $\bm{\rho}$ is to guarantee the non-negativity of standard deviation. Define $ \bm{v}=(\bm{\mu}, \bm{\rho})$. Then the random vector $\bm{\theta}$ is reparameterized as
$\bm{\theta}=h({\bm{v}},\bm{g})$, where
\begin{align} \label{eq: generation}
    \theta_m&=h({v_m}, g_m)  \\
    &=\mu_m+\log(1+\exp(\rho_m))\cdot g_m, ~~g_m \sim \mathcal{N}(0,1). \nonumber
\end{align}
For any $q \sim \{\mathcal{Q}^i\}_{i=1}^{N} \cup \mathcal{Q}^w$, the distribution of the random vector $\bm{\theta}$ is rewritten as $q_{\bm v} (\bm{\theta})$, where $q_{\bm v} (\theta_m)= \mathcal{N}\big{(}\mu_m,\log^2(1+\exp(\rho_m))\big{)}, m=1,\ldots,T$.



For the problem \eqref{Client-1} of the clients, we can get the closed-form results for KL divergence terms but cannot get that for the other term. In particular, we use Monte Carlo estimation to approximate this term. To speedup the convergence, we use the minibatch gradient descent (GD) algorithm. So the stochastic estimator for the $i$-th client is given by
\begin{multline}
\label{Client-FedBayes-2}
   \Omega^i(\bm{v}) \approx 
   - \frac{n}{b} \frac{1}{K} \sum_{j=1}^{b} \sum_{k=1}^{K}    \log p^i_{h({\bm{v}}, \bm{g}_k)}(\bm{D}^i_j) \\+  \zeta\, \mbox{KL}(q^i_{\bm{v}}(\bm{\theta})||w_{\bm{v}}(\bm{\theta})),
\end{multline}
where $b$ and $K$ are minibatch size and Monte Carlo sample size, respectively. 
Define the global model that has been locally updated but not yet aggregated on the server as {\em localized global model}. The cost function of localized global model for $i$-th client is represented by
\begin{align}
\label{Client-pFedBayes-4}
  \Omega^i_w(\bm{v}) = 
     \mbox{KL}(q^i_{\bm{v}}(\bm{\theta})||w_{\bm{v}}(\bm{\theta})).
\end{align}
At each iteration, the clients update their personalized models with $\nabla_{\vv} \Omega^i(\bm{v})$ and update the localized global model with $\nabla_{\vv} \Omega^i_w(\bm{v})$ alternatively. After $R$ iterations, the clients upload the localized global models to the server. On the server side, since clients are sometimes silent, we assume that a random set $\mathbb{S}^t \in \{1,\ldots,N\}$ with size $S$ is available for the server. After receiving the uploaded models, the server uses the average of clients in $\mathbb{S}^t $ to update the global model. Like \cite{t2020personalized,karimireddy2020scaffold}, an additional parameter $\beta$ is utilized to make the algorithm converge faster. The algorithm is shown in Algorithm \ref{Algorithm_PerFedBayes}.



\begin{table*}[!ht] 
\caption{Results on MNIST, FMNIST and CIFAR-10. Best results are bolded.}
  \label{dif-alg-table}
  \small
  \centering
  \scalebox{1}{
  \begin{tabular}{cccccccc}
    \toprule
    \multirow{2}{*}{Dataset}  & \multirow{2}{*}{Method} & \multicolumn{2}{c}{Small (Acc. ($\%$))}  & \multicolumn{2}{c}{Medium (Acc. ($\%$))}  & \multicolumn{2}{c}{Large (Acc. ($\%$))}\\
    \cmidrule(r){3-4}
    \cmidrule(r){5-6}
    \cmidrule(r){7-8}
        &   & {PM} & {GM}& {PM}  &{GM}& {PM}  &{GM}\\
    \midrule
    \multirow{8}{*}{MNIST} & FedAvg & - & 87.38$\pm$0.27 & - & 90.60$\pm$0.19 & - & 92.39$\pm$0.24\\
       & Fedprox  & - & 87.65$\pm$0.30 & - & 90.66$\pm$0.17 & - & 92.42$\pm$0.23\\
       & BNFed  & - & 78.70$\pm$0.69 & - & 80.02$\pm$0.60& -& 82.95$\pm$0.22 \\ 
       & Per-FedAvg & 89.29$\pm$0.59 & - & 95.19$\pm$0.33 & -& 98.27$\pm$0.08 & -\\
       & pFedMe & 92.88$\pm$0.04 & 87.35$\pm$0.08 & 95.31$\pm$0.17 & 89.67$\pm$0.34 & 96.42$\pm$0.08 & 91.25$\pm$0.14\\
       & HeurFedAMP & 90.89$\pm$0.17 & -& 94.74$\pm$0.07 & -& 96.90$\pm$0.12 & -\\
       & pFedGP & 85.96$\pm$2.30 & -& 91.96$\pm$0.97 & -& 95.66$\pm$0.43 & -\\
       & Ours     & \textbf{94.13$\pm$0.27} & \textbf{90.44$\pm$0.45} & \textbf{97.09$\pm$0.13} & \textbf{92.33$\pm$0.76} & \textbf{98.79$\pm$0.13} & \textbf{94.39$\pm$0.32} \\
    \midrule
    \multirow{8}{*}{FMNIST}
       & FedAvg & - & 81.51$\pm$0.19 & - & 83.90$\pm$0.13 & - & \textbf{85.42$\pm$0.14}\\
       & Fedprox  & - & \textbf{81.53$\pm$0.08} & - & \textbf{83.92$\pm$0.21} & - & 85.32$\pm$0.14\\
       & BNFed  & - & 66.54$\pm$0.64 & - & 69.68$\pm$0.39 & - & 70.10$\pm$0.24\\ 
       & Per-FedAvg & 79.79$\pm$0.83 & - & 84.90$\pm$0.47 & -& 88.51$\pm$0.28 & -\\
       & pFedMe  & 88.63$\pm$0.07 & 81.06$\pm$0.14 & 91.32$\pm$0.08 & 83.45$\pm$0.21 & 92.02$\pm$0.07 & 84.41$\pm$0.08 \\
       & HeurFedAMP  & 86.38$\pm$0.24 & -& 89.82$\pm$0.16 & -& 92.17$\pm$0.12 & -\\
       & pFedGP  & 86.99$\pm$0.41 & -& 90.53$\pm$0.35 & -& 92.22$\pm$0.13 & -\\
       & Ours     & \textbf{89.05$\pm$0.17} & 80.17$\pm$0.19 & \textbf{91.95$\pm$0.02} & 82.33$\pm$0.37 & \textbf{93.01$\pm$0.10} & 83.30$\pm$0.28 \\
    \midrule
    \multirow{8}{*}{CIFAR-10}
       & FedAvg & - & 44.24$\pm$3.01 & - & 56.73$\pm$1.81& - & \textbf{79.05$\pm$0.44}\\
       & Fedprox  & - & 43.70$\pm$1.38 & - & 57.35$\pm$3.11 & - & 77.65$\pm$1.62\\
       & BNFed & - & 34.00$\pm$0.16 & - & 39.52$\pm$0.56 & - & 44.37$\pm$0.19 \\ 
       & Per-FedAvg & 33.96$\pm$1.12 & - & 52.98$\pm$1.21 & -& 69.61$\pm$1.21 & -\\
       & pFedMe  & 49.66$\pm$1.53 & 43.67$\pm$2.14 & 66.75$\pm$1.87 & 51.18$\pm$2.57 & 77.13$\pm$1.06 & 70.86$\pm$1.04\\
       & HeurFedAMP  & 46.72$\pm$0.39 & -& 59.94$\pm$1.42 & -& 73.24$\pm$0.80 & -\\
       & pFedGP & 43.66$\pm$0.32 & -& 58.54$\pm$0.40 & -& 72.45$\pm$0.19 & -\\
       & Ours     & \textbf{61.37$\pm$1.40} & \textbf{47.71$\pm$1.19} & \textbf{73.94$\pm$0.97} & \textbf{60.84$\pm$1.26} & \textbf{83.46$\pm$0.13} & 64.40$\pm$1.22 \\
    \bottomrule
  \end{tabular}}
\end{table*}

\begin{figure*}[!ht]
    \centering
    \vspace{-12pt}
    \subfloat{\includegraphics[width=0.3\linewidth]{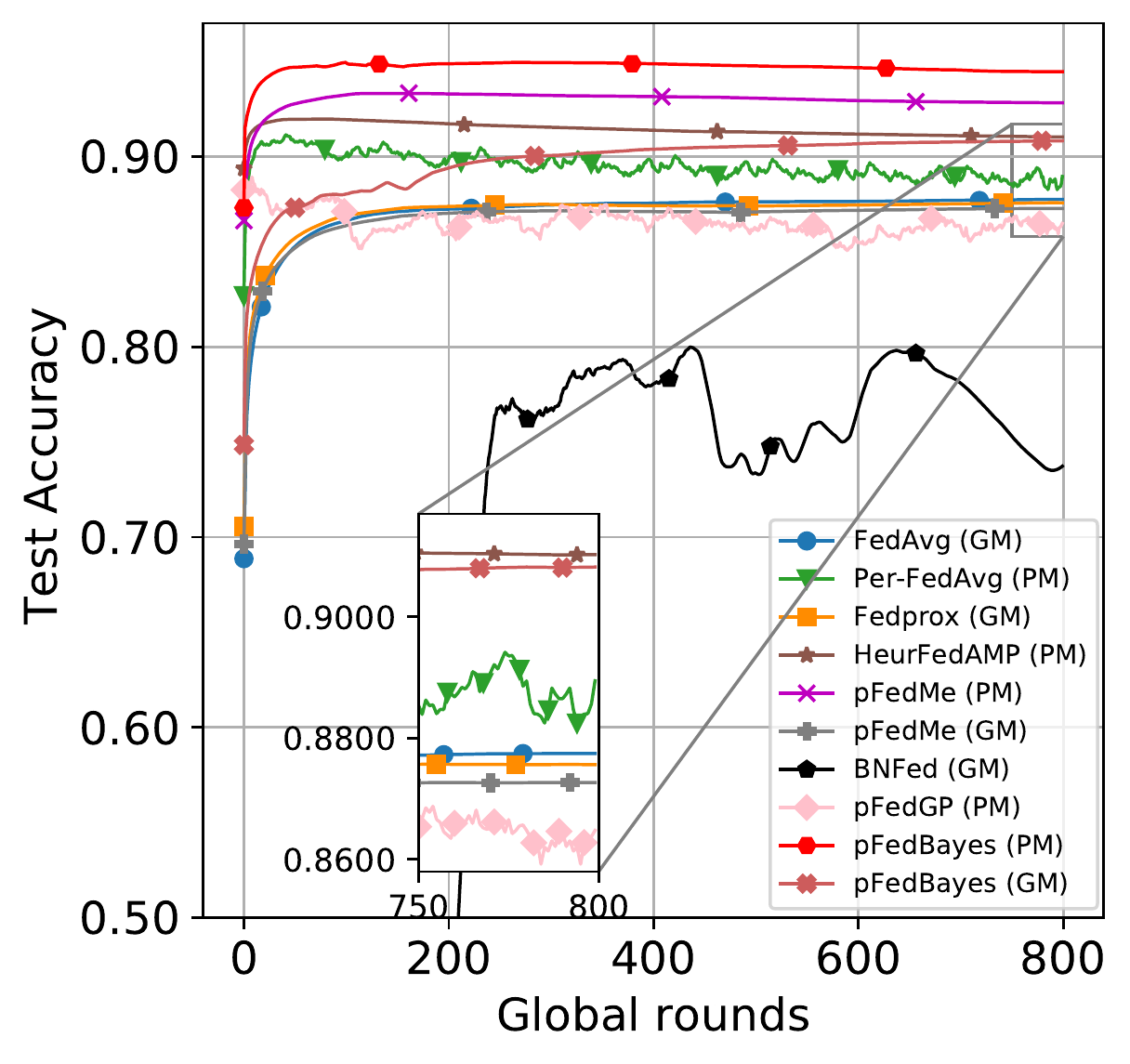}}
    \subfloat{\includegraphics[width=0.3\linewidth]{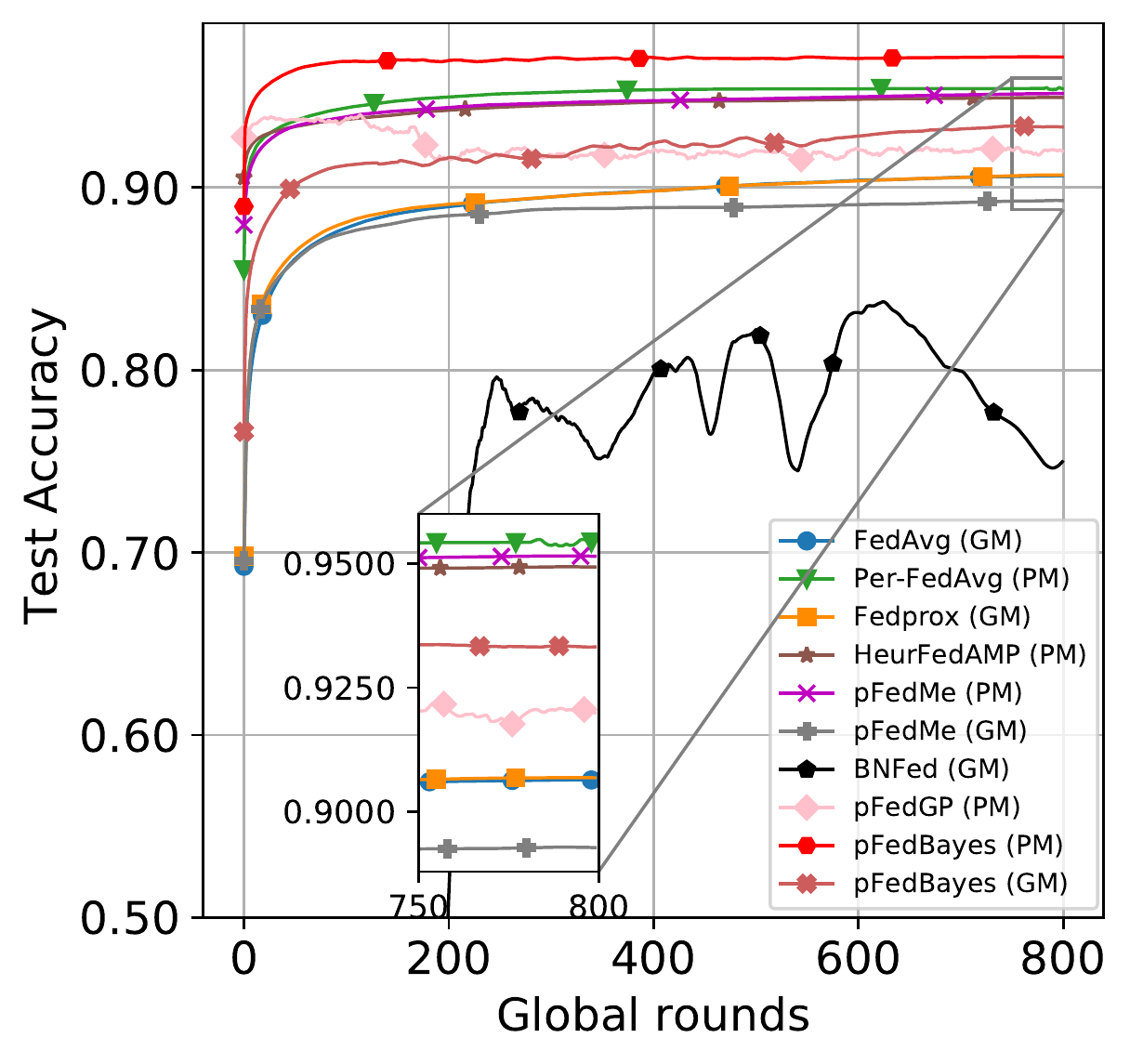}}
    \subfloat{\includegraphics[width=0.3\linewidth]{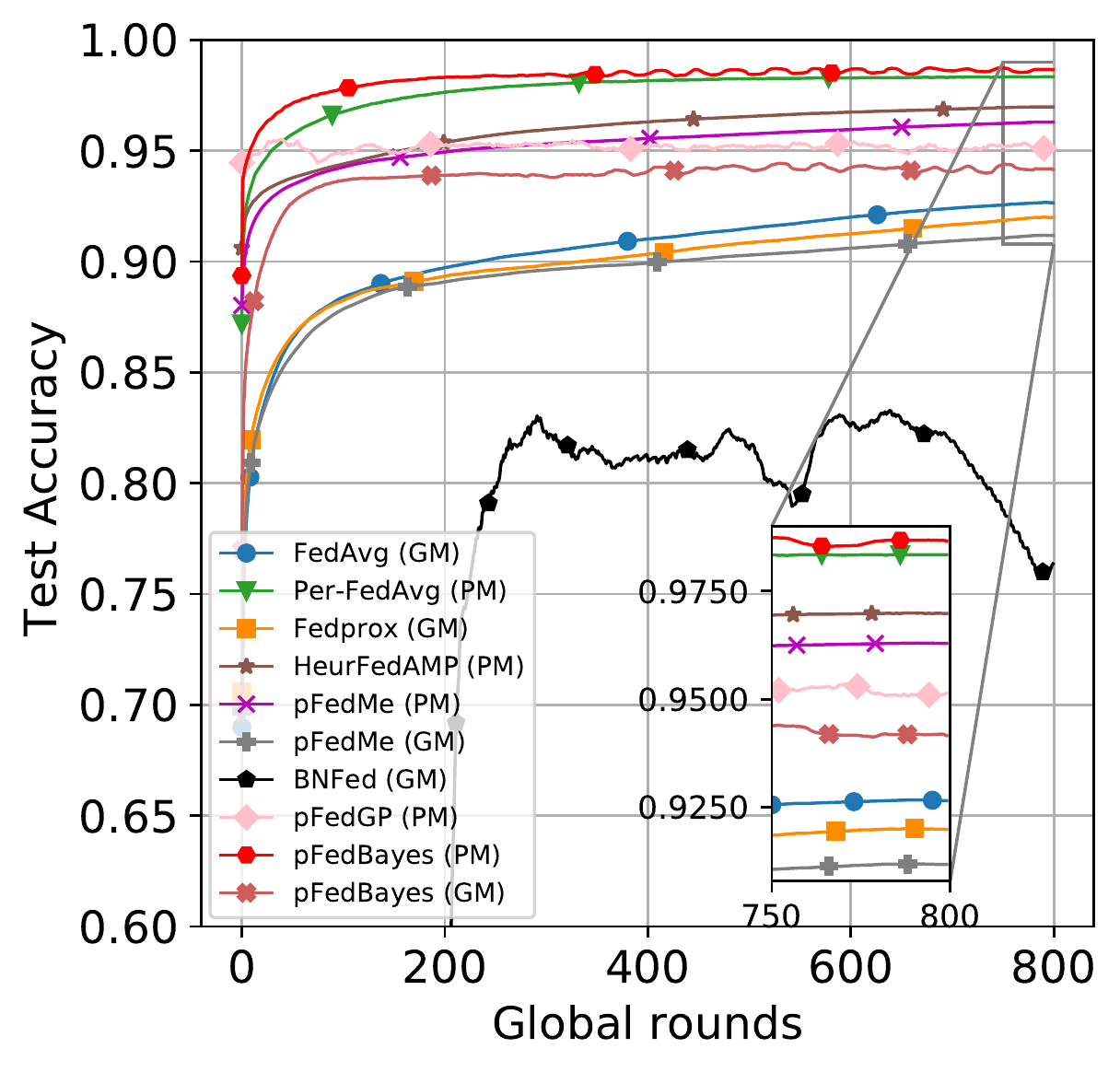}}
    
    
    \caption{Comparison results of the convergence rate of different algorithms on the MNIST dataset. \textbf{Left:} Small dataset. \textbf{Middle:} Medium dataset. \textbf{Right}: Large dataset.}
    \label{fig:fig3}
\end{figure*}

\section{Experiments}
\subsection{Experimental Setting}\label{sec_set}

We compare the performance of the proposed method pFedBayes with FedAvg \cite{mcmahan2017communication}, Fedprox \cite{li2018federated}, BNFed \cite{yurochkin2019bayesian}, Per-FedAvg \cite{fallah2020personalized}, pFedMe \cite{t2020personalized}, HeurFedAMP \cite{huang2021personalized} and pFedGP \cite{achituve2021personalized} based on non-i.i.d. datasets. We generate the non-i.i.d. datasets based on three public benchmark datasets, MNIST \cite{lecun2010mnist,lecun1998gradient}, FMNIST (Fashion-MNIST)\cite{xiao2017fashion} and CIFAR-10 \cite{krizhevsky2009learning}. For MNIST, FMNIST and CIFAR-10 datasets, we follow the non-i.i.d. setting strategy in \cite{t2020personalized}. Each client occupies a unique local data and only has 5 of the 10 labels. The number of clients for MNIST/FMNIST datasets is set as 10, while the number of clients for CIFAR-10 dataset is set as 20.

In addition, we split small, medium and large datasets on each public dataset to further validate the performance of our algorithm. For small, medium and large datasets of MNIST/FMNIST, there were 50, 200, 900 training samples and 950, 800, 300 test samples for each class, respectively. For the small, medium and large datasets of CIFAR-10, there were 25, 100, 450 training samples and 475, 400, 150 test samples in each class, respectively. For the random subset of clients $S$, we set $S=10$ for experiments on MNIST, FMNIST and CIFAR-10 datasets. We evaluate the performance of all algorithms by computing the highest accuracy over 700 to 800 communication rounds.




We did all experiments in this paper using servers with two GPUs (NVIDIA Tesla P100 with 16GB memory), two CPUs (each with 22 cores, Intel(R) Xeon(R) Gold 6152 CPU @ 2.10GHz), and 192 GB memory. The base DNN and VGG models and federated learning environment are implemented according to the settings in \cite{t2020personalized}. In particular, the DNN model uses one hidden layer, ReLU activation, and a softmax layer at the end. For the MNIST and FMNIST datasets, the size of the hidden layer is 100. The VGG model~\cite{simonyan2014very} is implemented for CIFAR-10 dataset with ``[16, `M', 32, `M', 64, `M', 128, `M', 128, `M']" cfg setting. We use PyTorch~\cite{NEURIPS2019_bdbca288} for all experiments.

\subsection{Experimental Hyperparameter Settings}

We first investigate the impact of hyperparameters on all algorithms based on the medium MNIST dataset and try to find the best hyperparameter settings for each algorithm to fairly compare their performance. For all hyperparameters, we refer to the settings in the corresponding papers of each algorithm and tune the parameters around their recommended values. For example, many algorithms use a learning rate $\eta = 0.005$ (e.g., pFedMe),  then we set the learning rate in the range of 0.0005 to 0.1 for parameter tuning. For some algorithm-specific hyperparameters, we use the values recommended in their papers. More detailed results of parameter tuning are given in Appendix~B.

Based on the experimental results, we set the learning rate of FedAvg and Per-FedAvg to 0.01. The learning rate and regularization weight of Fedprox are respectively set as 0.01 and $\lambda = 0.001$. The learning rate of BNFed is set as 0.5, while for pFedGP is set as 0.05. The personalized learning rate, global learning rate and regularization weight of pFedMe are respectively set as 0.01, 0.01 and $\lambda = 15$. The learning rate and regularization weight of HeurFedAMP are respectively set as 0.01 and $\alpha = 5$. For the proposed {\texttt{pFedBayes}}, we set the initialization of weight parameters $\rho = -2.5$, the tradeoff parameter $\zeta = 10$, and the learning rates of the personalized model and global model $\eta_1 = \eta_2 = 0.001$.

\subsection{Performance Comparison Results}



Table 1 shows the performance of each algorithm under the optimal hyperparameter settings. On small, medium and large datasets of MNIST, the personalized model of our algorithm outperforms other SOTA comparison algorithms by 1.25$\%$, 1.78$\%$ and 0.52$\%$, while the global model outperforms other SOTA models by 2.79$\%$, 1.67$\%$ and 1.97$\%$, respectively. We can see that our algorithm performs better when the amount of data is small, thanks to the advantage of the Bayesian algorithm on small samples.
Figure 2 shows the comparison results of the convergence speed of different algorithms on the MNIST dataset. We can see that our {\texttt{pFedBayes}} converges significantly under a small dataset, and the performance is stable after 50 iterations, clearly ahead of other algorithms.


On the small, medium and large datasets of FMNIST, the personalization model of our algorithm outperforms other SOTA comparison algorithms by 0.42$\%$, 0.63$\%$ and 0.79$\%$, respectively. However, the accuracy of the global model is not ahead. This is mainly because the FMNIST dataset is relatively simple and does not have good non-i.i.d. features. Therefore, the advantages of personalization algorithms are not obvious. This can be verified by the fact that FedAvg achieves optimal performance on large datasets.

On the small, medium and large datasets of CIFAR-10, the personalization model of our algorithm respectively outperforms other SOTA comparison algorithms by 11.71$\%$. 7.19$\%$ and 6.33$\%$. It can be seen that for complex non-i.i.d. dataset, our {\texttt{pFedBayes}} can achieve substantial improvement on small samples. The global model also outperforms other SOTA models by 3.47$\%$ and 3.49$\%$ on small and medium datasets, respectively. But on large datasets, our global model has no performance advantage. This is because the excellent performance of pFedBayes under small samples is attributed to BNN \cite{blundell2015weight,khan2019deep}. For large datasets, BNNs need other techniques to achieve the same performance as non-Bayesian algorithms~\cite{Osawa2019Practical}. For fairness, we use the same techniques in large datasets as that in small datasets, resulting in performance degradation when aggregating models.


Furthermore, our {\texttt{pFedBayes}} can provide results for model uncertainty estimation, a useful information in federated learning. See Appendix~B.2 for more detailed results.



\section{Conclusions}

In this paper, we propose a novel personalized federated learning model via variational Bayesian inference. Each client uses the aggregated global distribution as prior distribution and updates its personal distribution by balancing the construction error over its personal data and the KL divergence with aggregated global distribution. We provide theoretical analysis for the averaged generalization error of all clients, which shows that the proposed model achieves minimax-rate optimality up to a logarithmic factor. An efficient algorithm is given by updating the global model and personal models alternatively. Extensive experiments present that the proposed algorithm outperforms many advanced personalization methods in most cases, especially when the amount of data is limited.

\section*{Acknowledgements}

This work is sponsored by China National Postdoctoral Program for Innovative Talents under Grant No.
BX2021346.


\newpage

\onecolumn
\appendix
\setcounter{equation}{0}
\renewcommand{\theequation}{A.\arabic{equation}}

\setcounter{theorem}{0}
\renewcommand{\thetheorem}{A\arabic{theorem}}

\setcounter{lemma}{0}
\renewcommand{\thelemma}{A.\arabic{lemma}}

\appendix

\section{Proof of Lemmas}

\subsection{Proof of Lemma \ref{lm: upper bound 1}}

Before prove Lemma \ref{lm: upper bound 1}, let's review an important lemma. Lemma \ref{lmdonsker} restates the  Donsker and Varadhan's representation for the $\mbox{KL}$ divergence, whose proof can be found in \cite{Boucheron2013Concentration}. 
\begin{lemma}{\label{lmdonsker}}
	For any probability measure $\mu$ and any measurable function $h$ with $e^h \in L_1(\mu)$,
	$$
	\log \int e^{h(\eta)}\mu(d\eta) = \sup_{\rho}\left[\int h(\eta)\rho(d \eta) - \mbox{\rm KL}(\rho||\mu)\right].
	$$
\end{lemma}

Now we are ready to prove Lemma \ref{lm: upper bound 1}. 
Let $\log\eta(P^{i}_{\bm{\theta}}, P^{i}) = l_n(P^{i}_{\bm{\theta}}, P^{i})/\zeta+ n \, d^2(P^{i}_{\bm{\theta}}, P^{i})$. Since $\zeta\ge 1$, using Jensen's inequality and the concavity of $(\cdot)^{1/\zeta}$ gets 
\begin{equation}
    \mathbb{E}_{P^i} \exp(l_n(P^{i}_{\bm{\theta}}, P^{i})/\zeta)= 
    \mathbb{E}_{P^i} \left(\frac{p^{i}_{\bm{\theta}}(\bm{D}^i)}{p^{i}(\bm{D}^i)} \right)^{\frac{1}{\zeta}} \le  \left(\mathbb{E}_{P^i} \frac{p^{i}_{\bm{\theta}}(\bm{D}^i)}{p^{i}(\bm{D}^i)} \right)^{\frac{1}{\zeta}}=1.
\end{equation}
Together with the proof from the Theorem 3.1 of \cite{pati2018statistical}, we have 
	\begin{equation}
	\int_{\Theta}\eta(P^{i}_{\bm{\theta}}, P^{i}) w^\star(\bm{\theta})  d\bm{\theta} \leq e^{C n\varepsilon^{2}_{n}},\, \mbox{w.h.p.,}
	\end{equation}
where $C>0$ is a large constant.

By using Lemma \ref{lmdonsker} with $h(\eta)=\log\eta(P^{i}_{\bm{\theta}},P^i)$, $\mu=w^\star(\bm{\theta})$ and $\rho=\hat{q}^i(\bm{\theta})$, we obtain 
\begin{align} \label{neq: local_upperbound1}
  \int_{\Theta}  d^2(P_{\bm{\theta}}^i, P^i) \hat{q}^i(\bm{\theta}) d\bm{\theta} 
  \le &  \frac{1}{n}\left[ \frac{1}{\zeta}\int_{\Theta} l_n( P^{i}, P^{i}_{\bm{\theta}})\hat{q}^i(\bm{\theta}) d\bm{\theta} +\mbox{KL}(\hat{q}^i(\bm{\theta})||w^\star(\bm{\theta})) + \log \int_{\Theta} \eta(P^{i}_{\bm{\theta}}, P^{i}) w^\star(\bm{\theta})  d\bm{\theta} \right]
  \nonumber\\
  \le&  \frac{1}{n}\left[ \frac{1}{\zeta} \int_{\Theta} l_n( P^{i}, P^{i}_{\bm{\theta}})\hat{q}^i(\bm{\theta}) d\bm{\theta} + \mbox{KL}(\hat{q}^i(\bm{\theta})||w^\star(\bm{\theta})) \right]+ C\varepsilon^{2}_{n}. 
\end{align}

By averaging over all $N$ clients, we have 
\begin{multline} \label{neq: local_upperbound_sum_a}
    \frac{1}{N} \sum_{i=1}^{N} \int_{\Theta}  d^2(P_{\bm{\theta}}^i, P^i) \hat{q}^i(\bm{\theta}) d\bm{\theta} \le 
    \frac{1}{n} \left\{\frac{1}{N} \sum_{i=1}^{N} \left[ \frac{1}{\zeta} \int_{\Theta} l_n(P^{i}, P^{i}_{\bm{\theta}})\hat{q}^i(\bm{\theta}) d\bm{\theta} + \mbox{KL}(\hat{q}^i(\bm{\theta})||w^\star(\bm{\theta})) \right] \right\} + C\varepsilon^{2}_{n}.
\end{multline}


\subsection{Proof of Lemma \ref{lm: upper bound 2}}

Before prove Lemma \ref{lm: upper bound 2},  we present a lemma to give the optimal $w(\bm{\theta})$ for given distribution of each client $q^i(\bm{\theta})$, which is proved in Section \ref{Proof_Lemma_A2}. 

\begin{lemma} \label{lm: optimalsolution}
Let $w(\bm{\theta})$ be the optimal variational solution of the problem 
\begin{equation} \label{eq: F_w_m}
   \min_{w(\bm{\theta})} \left\{ F(w)=\frac{1}{N} \sum_{i=1}^{N} \mbox{\rm KL}(q^i(\bm{\theta})||w(\bm{\theta})))\right\}.
\end{equation}
Then we have
\begin{align}
    \mu_{w,m}&=\frac{1}{N} \sum_{i=1}^{N} \mu_{i,m},\\
    \sigma_{w,m}^2&=\frac{1}{N} \sum_{i=1}^{N} \left[\sigma_{i,m}^2+\left(\mu_{i,m}-\mu_{w,m}\right)^2\right] =\frac{1}{N} \sum_{i=1}^{N} \left[\sigma_{i,m}^2+\mu_{i,m}^2-\mu_{w,m}^2\right].
\end{align}
\end{lemma}

Now we are ready to prove Lemma \ref{lm: upper bound 2}. Choosing $\bm{\theta}^\star_i$ that minimizes $\norm{f^i_{\bm{\theta}}-f^i}^2_\infty$ subject to $\norm{\bm{\theta}}_\infty \le B$, then we define $\tilde{q}^i(\bm{\theta})$ as follows, for $m=1,\dots, T$:
\begin{equation}\label{vbstar}
	\begin{split}
	& \theta_{i,m} \sim \mathcal{N}(\theta_{i,m}^\star,\sigma^2_{n}),
	\end{split}
\end{equation}
where
\begin{equation} \label{eq: sigma_i_n}
    \sigma_{n}^2=\frac{T}{8n} A
\end{equation}
and 
\begin{equation} \label{eq: A2}
A=\log^{-1}(3s_0M)\cdot(2BM)^{-2(L+1)}\left[\left(s_0 + 1 + \frac{1}{BM-1}\right)^2 + \frac{1}{(2BM)^2-1}
	+\frac{2}{(2BM-1)^2}\right]^{-1}.
\end{equation}

Let $\tilde{w}(\bm{\theta})$ be the optimal solution for 
\begin{equation}
    \min_{w(\bm{\theta}) \in \mathcal{Q}}\frac{1}{N} \sum_{i=1}^{N} \mbox{KL}(\tilde{q}^i(\bm{\theta})||w(\bm{\theta}))).
\end{equation}
Then by using Lemma \ref{lm: optimalsolution}, the distribution $\tilde{w}(\bm{\theta})$ satisfies
\begin{equation}\label{wtilde}
\theta_{\tilde{w},m} \sim \mathcal{N}(\mu_{\tilde{w},m}, \sigma^2_{\tilde{w},m}), m=1,\ldots, T,
\end{equation}
where
\begin{align}
    \mu_{\tilde{w},m}&=\frac{1}{N} \sum_{i=1}^{N} \theta_{i,m}^\star,\\
    \sigma_{\tilde{w},m}^2&=\frac{1}{N} \sum_{i=1}^{N} \left[\sigma_{n}^2+\left(\theta_{i,m}^\star-\mu_{\tilde{w},m}\right)^2\right]=\sigma_{n}^2-\mu_{\tilde{w},m}^2+\frac{1}{N} \sum_{i=1}^{N} \theta_{i,m}^{\star2}. \label{eq: sigma_w_tilde}
\end{align}
 
 Since $w^\star(\bm{\theta})$ and $\hat{q}(\bm{\theta})$ correspond to the optimal solution of the global problem, then we obtain
 \begin{multline}
    \frac{1}{N} \sum_{i=1}^{N}  \left[  \int_{\Theta} l_n(P^{i},P^{i}_{\bm{\theta}}) \hat{q}^i(\bm{\theta}) d\bm{\theta} +  \zeta\, \mbox{KL}(\hat{q}^i(\bm{\theta})||w^\star(\bm{\theta}))\right]
    \\ \le
     \frac{1}{N} \sum_{i=1}^{N}  \left[  \int_{\Theta} l_n(P^{i},P^{i}_{\bm{\theta}}) \tilde{q}^i(\bm{\theta}) d\bm{\theta} +  \zeta\, \mbox{KL}(\tilde{q}^i(\bm{\theta})||\tilde{w}(\bm{\theta}))\right].
 \end{multline}

Next we will give the upper bound under probability distributions $\tilde{w}(\bm{\theta})$ and $\tilde{q}^i(\bm{\theta})$. According to the definitions above and the mean-field decomposition, we can represent the probabilities as follows
	\begin{equation}
	\tilde{q}^i(\bm{\theta}) = \prod^{T}_{m=1}\mathcal{N}(\theta^\star_{i,m}, \sigma^2_{n}),
	\end{equation}
	\begin{equation} \label{eq: sigma_tilde 2}
	\tilde{w}(\bm{\theta}) =\prod^T_{m=1}\mathcal{N}(\mu_{\tilde{w},m}, \sigma^2_{\tilde{w},m}).
	\end{equation}

	Then we have
	\begin{align} \label{eq:KL_q_w}
	\mbox{KL}(\tilde{q}^i(\bm{\theta})||\tilde{w}(\bm{\theta}))
 	= &\mbox{KL}\left(\prod^{T}_{m=1}\mathcal{N}(\theta^\star_{i,m}, \sigma^2_{n}) \Bigl|\Bigr| \prod^T_{m=1}\mathcal{N}(\mu_{\tilde{w},m}, \sigma^2_{\tilde{w},m})\right) \nonumber\\
	= &  \sum^T_{m=1}\mbox{KL}\left(\mathcal{N}(\theta^\star_{i,m}, \sigma^2_{n})|| \mathcal{N}(\mu_{\tilde{w},m}, \sigma^2_{\tilde{w},m}) \right) \nonumber\\
	= &  \frac{1}{2}\sum^T_{m=1}\left[\log\Bigl(\frac{\sigma^2_{\tilde{w},m}}{\sigma^2_{n}}\Bigr)+\frac{\sigma^2_{n}+(\theta^\star_{i,m}-\mu_{\tilde{w},m})^{2}}{\sigma^2_{\tilde{w},m}} - 1\right] \nonumber\\
    = & \frac{1}{2}\sum^T_{m=1}\left[\log\Bigl(\frac{\sigma^2_{\tilde{w},m}}{\sigma^2_{n}}\Bigr) - 1\right]+\frac{1}{2} \sum^T_{m=1} \frac{\sigma^2_{n}+(\theta^\star_{i,m}-\mu_{\tilde{w},m})^{2}}{\sigma^2_{\tilde{w},m}} \nonumber\\
    \leq &  \frac{T}{2}\left[\log\left(\frac{\sigma^2_{n}+B^2}{\sigma^2_{n}}\right) - 1\right]+\frac{1}{2} \sum^T_{m=1} \frac{\sigma^2_{n}+(\theta^\star_{i,m}-\mu_{\tilde{w},m})^{2}}{\sigma^2_{\tilde{w},m}},
	\end{align}
 where the inequality applies 
\begin{equation*}
    \sigma_{\tilde{w},m}^2= \sigma_{n}^2-\mu_{\tilde{w},m}^2+\frac{1}{N} \sum_{i=1}^{N} \theta_{i,m}^{\star2}\le \sigma^2_{n}+B^2.
\end{equation*}

So the sum of the KL divergence (\ref{eq:KL_q_w}) satisfies
\begin{align*}
	\frac{1}{N}\sum_{i=1}^{N}\mbox{KL}(\tilde{q}^i(\bm{\theta})||\tilde{w}(\bm{\theta}))
     &\le  \frac{1}{N}\sum_{i=1}^{N} \left\{ \frac{T}{2}\left[\log\left(\frac{\sigma^2_{n}+B^2}{\sigma^2_{n}}\right) - 1\right]+\frac{1}{2} \sum^T_{m=1} \frac{\sigma^2_{n}+(\theta^\star_{i,m}-\mu_{\tilde{w},m})^{2}}{\sigma^2_{\tilde{w},m}} \right\}\\
     &\leq   \frac{T}{2}\log\left(\frac{\sigma^2_{n}+B^2}{\sigma^2_{n}}\right),
\end{align*}
where the last inequality applies (\ref{eq: sigma_w_tilde}) such that 
\begin{equation*}
    \frac{1}{N}\sum_{i=1}^{N} \frac{\sigma^2_{n}+(\theta^\star_{i,m}-\mu_{\tilde{w},m})^{2}}{\sigma^2_{\tilde{w},m}}=1.
\end{equation*}

Under Assumption 3, we obtain
\begin{align} \label{eq: sum_kl_q_w_2}
	\frac{1}{N}\sum_{i=1}^{N}\mbox{KL}(\tilde{q}^i(\bm{\theta})||\tilde{w}(\bm{\theta})) 
     \leq  \frac{T}{2}  \log \left(\frac{2B^2}{\sigma^2_{n}}\right).
\end{align}


Incorporating the definition of $\sigma_n$ yields the following result
\begin{align*}
   \frac{T}{2}\log \left(\frac{2B^2}{\sigma^2_{n}}\right)
  \le T(L+1)\log(2BM)+\frac{T}{2}\log\log(3s_0 M)+T\log\left(4s_0 \sqrt{\frac{n}{T}}\right)+ \frac{T}{2} \log(2B^2) \le C' n r_n.
\end{align*}

Therefore, we get
\begin{align} \label{eq: sum_kl_q_w_3}
	\frac{1}{N}\sum_{i=1}^{N}\mbox{KL}(\tilde{q}^i(\bm{\theta})||\tilde{w}(\bm{\theta})) 
     \leq  C' n r_n.
\end{align}

By following the technique from the Supplementary of \cite{bai2020efficient}, we can give the following upper bound
\begin{align}
    \int_{\Theta} l_n(P^{i},P^{i}_{\bm{\theta}}) \tilde{q}^i(\bm{\theta}) d\bm{\theta}
    \le C''(nr_n+ n \xi^i_n).
\end{align}

Combining the above results and $\zeta \ge 1$ gives that 
\begin{equation}
    \frac{1}{N} \sum_{i=1}^{N}  \left[  \int_{\Theta} l_n(P^{i},P^{i}_{\bm{\theta}}) \hat{q}^i(\bm{\theta}) d\bm{\theta} +  \zeta\, \mbox{KL}(\hat{q}^i(\bm{\theta})||w^\star(\bm{\theta}))\right]  \le n\left(  C' \zeta r_n +\frac{C''}{N} \sum_{i=1}^{N}\xi^i_n \right).
 \end{equation}
 



\subsection{Proof of Lemma A.2} \label{Proof_Lemma_A2}

Since $w(\bm{\theta})$ is the optimal solution of Eq. \eqref{eq: F_w_m}, the partial derivatives of $F(w)$ with respect to $\mu_{w,m}$ and $\sigma_{w,m}$ are zero, i.e., 
\begin{align} 
    \frac{1}{N} \sum_{i=1}^{N} \frac{\partial \, \mbox{KL}(q^i(\bm{\theta})||w(\bm{\theta}))}{\partial \,\mu_{w,m}} &=0, \label{eq: pd1}\\
    \frac{1}{N} \sum_{i=1}^{N} \frac{\partial \, \mbox{KL}(q^i(\bm{\theta})||w(\bm{\theta}))}{\partial \,\sigma_{w,m}} &=0. \label{eq: pd2}
\end{align}

According to Eq. \eqref{kl_q_w}, we can get the partial derivatives 
\begin{align}
    \frac{1}{N} \sum_{i=1}^{N} \frac{\partial \, \mbox{KL}(q^i(\bm{\theta})||w(\bm{\theta}))}{\partial \,\mu_{w,m}} &= \frac{1}{N} \sum_{i=1}^{N} \frac{-2(\mu_{i,m}-\mu_{w,m})}{\sigma_{w,m}^2}, \label{eq: pd1a}\\
    \frac{1}{N} \sum_{i=1}^{N} \frac{\partial \, \mbox{KL}(q^i(\bm{\theta})||w(\bm{\theta}))}{\partial \,\sigma_{w,m}} &= \frac{1}{N} \sum_{i=1}^{N} \left[ \frac{2}{\sigma_{w,m}}- \frac{2 \left[\sigma_{i,m}^2+\left(\mu_{i,m}-\mu_{w,m}\right)^2\right]}{\sigma_{w,m}^3} \right].\label{eq: pd2a}
\end{align}

Combining Eqs. (\ref{eq: pd1}), (\ref{eq: pd2}), (\ref{eq: pd1a}) and (\ref{eq: pd2a})  yields 
\begin{align}
    \mu_{w,m}&=\frac{1}{N} \sum_{i=1}^{N} \mu_{i,m},\\
    \sigma_{w,m}^2&=\frac{1}{N} \sum_{i=1}^{N} \left[\sigma_{i,m}^2+\left(\mu_{i,m}-\mu_{w,m}\right)^2 \right]\nonumber\\
    &=\frac{1}{N} \sum_{i=1}^{N} \left[\sigma_{i,m}^2+\mu_{i,m}^2-\mu_{w,m}^2\right].
\end{align}

\section{Experimental Results on MNIST Dataset}

\subsection{Effect of Hyperparameters}\label{sec_result_hyper}

We test {\texttt{pFedBayes}} in a basic DNN model with 3 full connection layers [784, 100, 10] on the MNIST dataset. The results are listed in Table~\ref{detail-result-dnn}.
\textbf{Effects of $\eta_1$ and $\eta_2$:} In {\texttt{pFedBayes}} algorithm, $\eta_1$ and $\eta_2$ are respectively denote the learning rate of the personalized model and global model. We tune the learning rate in the range of $[0.001, 0.005]$ while fix other parameters. From Table~\ref{detail-result-dnn} we can see that $\eta_1 = \eta_2 = 0.001$ is the best setting. \textbf{Effects of $\zeta$:} In {\texttt{pFedBayes}} algorithm, $\zeta$ can adjust the degree of personalization of personalized models. Increasing $\zeta$ can improve the test accuracy of the global model and weaken the performance of the personalized model. On the basis of the best learning rate, we tune $\zeta \in \{0.5, 1, 5, 10, 20\}$. Table~\ref{detail-result-dnn} shows that $\zeta=10$ is the best setting. Hence, we set $\zeta = 10$ for the remaining experiments. \textbf{Effects of $\rho$:} It is known that the initialization of the weight parameters affects the results of the model. Hence, we also tune $\rho \in \{-1, -1.5, -2, -2.5, -3\}$. Table~\ref{detail-result-dnn} shows that $\rho=-2.5$ is the best setting. Hence, we set $\rho=-2.5$ for the remaining experiments.

The results of FedAvg, Fedprox, NBFed, Per-FedAvg, pFedMe, HeurFedAMP and pFedGP are listed in Table~\ref{detail-result-dnn1}. {The parameters $\eta_1$ and $\eta_2$ are the learning rates of the personalized model and the global model, respectively. ``Personalized'' means this column indicates whether this is a personalized algorithm.} In order to make a fair comparison with these algorithms, we perform the following hyperparameter adjustment. In \cite{t2020personalized}, $\lambda=15$ achieves the best performance and is the recommended setting, so we use this setting directly in our experiments. We tune $\lambda \in \{0.001, 0.01, 0.1, 1\}$ the same as the setting in Fedprox \cite{li2018federated}. For HeurFedAMP , we tune $\alpha \in [0.1, 0.5]$, and $0.5$ is the recommended setting in its paper \cite{huang2021personalized}. For pFedGP, the recommended learning rate in its paper is 0.05, we hence tune $\eta \in [0.01, 0.05, 0.1]$. We can see that $\eta = 0.05$ can achieve better performance. For other hyperparameters in pFedGP, we set them the same as in \cite{achituve2021personalized}. Notably, $\alpha$ represents the proportion of the client model that does not interact with the global model. The higher the value, the less interaction with the global model. For other algorithms, there are 10 client models interacting with the global model. For a fair comparison, the corresponding $\alpha$ should be set to 0.1. Clearly 0.5 provides better performance of the personalized model, although a setting of 0.5 is not a fair comparison. We still use the parameter setting of 0.5 recommended in its paper. In addition, it should be noted that the learning rate cannot be set too large, and an appropriate value should be taken. For most federated learning algorithms, a learning rate that is too large will cause the model to diverge, resulting in a severe loss of model aggregation performance. {From the experimental results of the personalized algorithms, we find that the optimal learning rate is between 0.0005 and 0.01. For a fair comparison, we also set the learning rates of the non-personalized algorithms, i.e. FedAvg and FedProx, within this range.}

\begin{table*}[!htb]
  \caption{Results of pFedBayes on Medium dataset (MNIST).}
  \label{detail-result-dnn}
  \centering
  \scalebox{1}{
  \begin{tabular}{cccccc}
    \toprule
    \multirow{1}{*}{$\rho$} & $\zeta$ & $\eta_1$ & $\eta_2$ & PM Acc.($\%$)& GM Acc.($\%$)\\
    \toprule
        \multirow{3}{*}
        -1 & 10 & 0.001 & 0.001 & 97.38 & 91.61\\
        -1 & 10 & 0.001 & 0.005 & 97.12 & 90.32\\
        -1 & 10 & 0.005 & 0.001 & 96.73 & 91.34\\
        -1 & 10 & 0.005 & 0.005 & 96.78 & 91.10\\
    \midrule
        \multirow{3}{*}
        -2 & 10 & 0.001 & 0.001 & 97.45 & 92.40\\
        -2 & 10 & 0.001 & 0.005 & 97.42 & 91.35\\
        -2 & 10 & 0.005 & 0.001 & 97.36 & 91.27\\
        -2 & 10 & 0.005 & 0.005 & 97.28 & 90.01\\
    \midrule
        \multirow{3}{*}
        -3 & 10 & 0.001 & 0.001 & 97.08 & 92.16\\
        -3 & 10 & 0.001 & 0.005 & 96.35 & 90.47\\
        -3 & 10 & 0.005 & 0.001 & 96.64 & 87.34\\
        -3 & 10 & 0.005 & 0.005 & 97.28 & 90.22\\
    \midrule
        \multirow{3}{*}
        -1.5 & 10 & 0.001 & 0.001 & 97.38 & 92.31\\
        -2.0 & 10 & 0.001 & 0.001 & 97.45 & 92.40\\
        -2.5 & 10 & 0.001 & 0.001 & 97.18 & 93.22\\
    \midrule
        \multirow{5}{*}
        -2.5 & 0.5 & 0.001 & 0.001 & 97.13 & 89.88\\
        -2.5 & 1 & 0.001 & 0.001 & 97.41 & 91.24\\
        -2.5 & 5 & 0.001 & 0.001 & 97.38 & 93.03\\
        \textbf{-2.5} & \textbf{10} & \textbf{0.001} & \textbf{0.001} & \textbf{97.18} & \textbf{93.22}\\
        -2.5 & 20 & 0.001 & 0.001 & 97.04 & 92.77\\
    \bottomrule
  \end{tabular}}
\end{table*}

\begin{figure*}[!htb]
    \centering
    \vspace{-12pt}
    \subfloat{\includegraphics[width=0.2\linewidth]{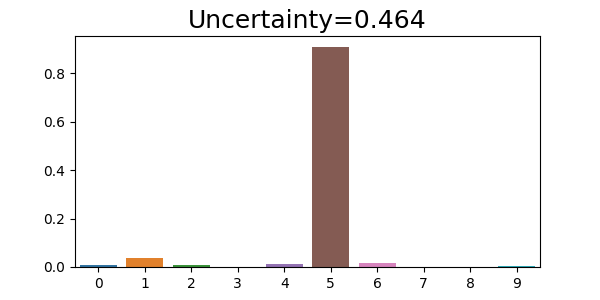}}
    \subfloat{\includegraphics[width=0.2\linewidth]{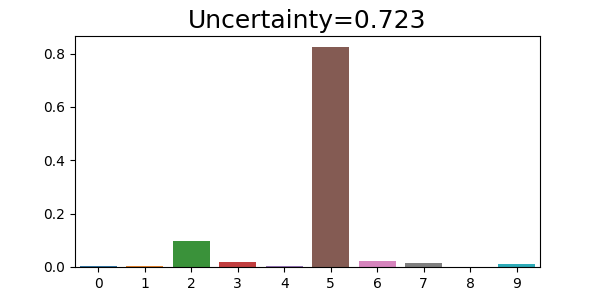}}
    \subfloat{\includegraphics[width=0.2\linewidth]{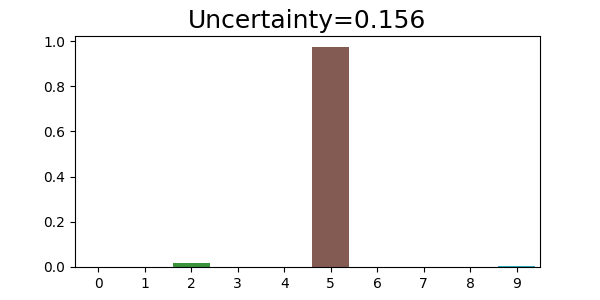}}
    \subfloat{\includegraphics[width=0.2\linewidth]{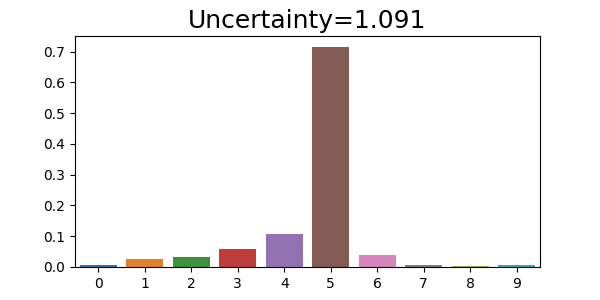}}
    \subfloat{\includegraphics[width=0.2\linewidth]{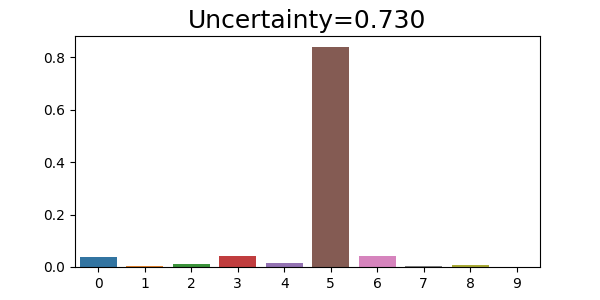}}
    
    \vspace{-12pt}
    \subfloat{\includegraphics[width=0.2\linewidth]{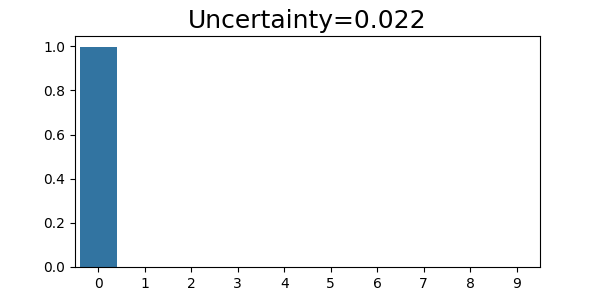}}
    \subfloat{\includegraphics[width=0.2\linewidth]{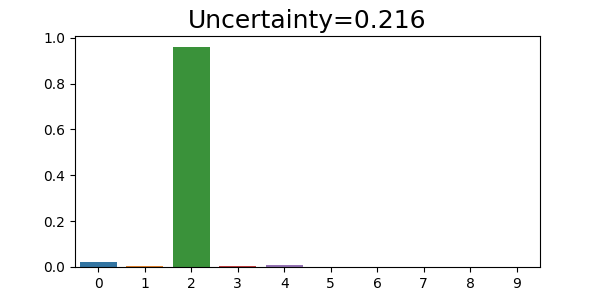}}
    \subfloat{\includegraphics[width=0.2\linewidth]{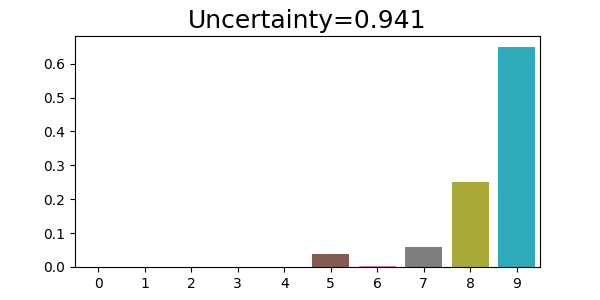}}
    \subfloat{\includegraphics[width=0.2\linewidth]{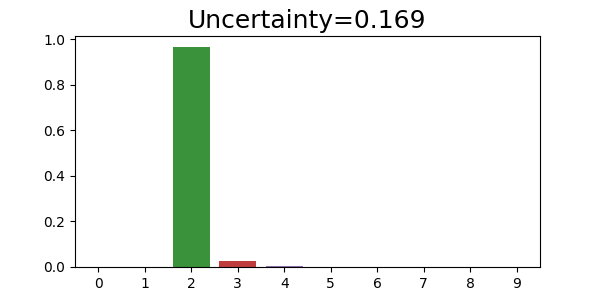}}
    \subfloat{\includegraphics[width=0.2\linewidth]{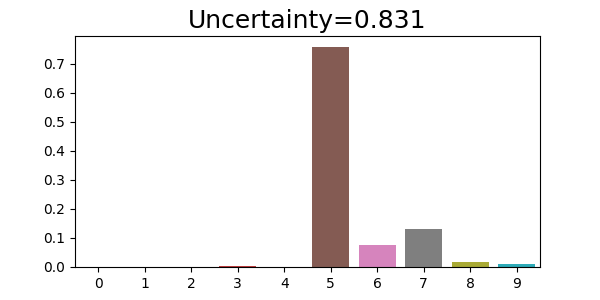}}
    
    \vspace{-12pt}
    \subfloat{\includegraphics[width=0.2\linewidth]{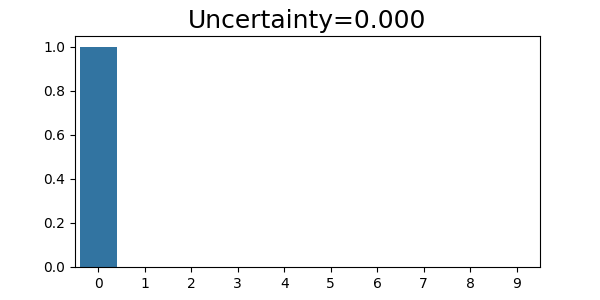}}
    \subfloat{\includegraphics[width=0.2\linewidth]{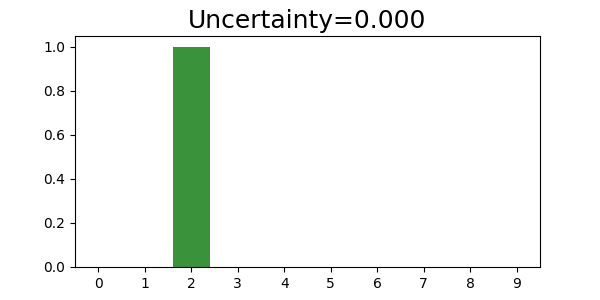}}
    \subfloat{\includegraphics[width=0.2\linewidth]{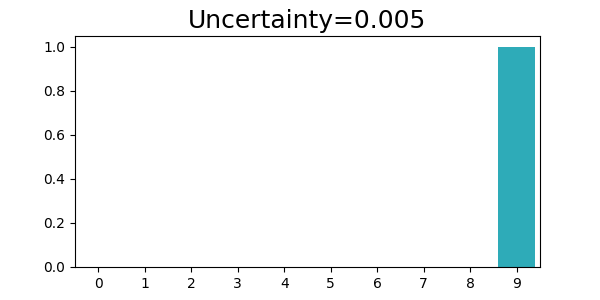}}
    \subfloat{\includegraphics[width=0.2\linewidth]{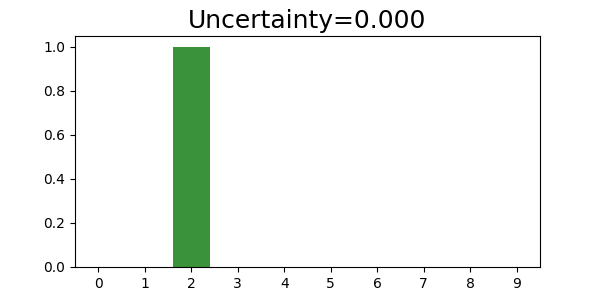}}
    \subfloat{\includegraphics[width=0.2\linewidth]{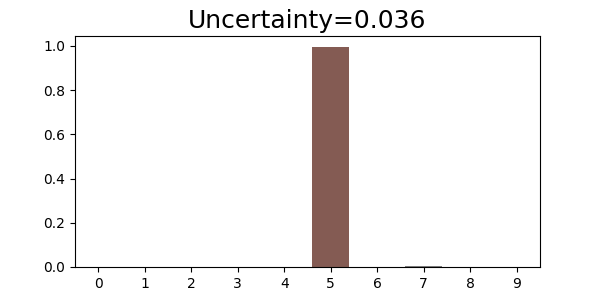}}
    
    \caption{Predication uncertainty of personalized Bayesian federated learning model with Gaussian assumptions. From the first row to the third row, the number of training rounds is 0, 1, and 10, respectively. The five columns represent five different clients.}
    \label{fig:fig2}
\end{figure*}

\begin{table*}[htbp]
  \caption{Comparison algorithm results on the medium dataset (MNIST).}
  \label{detail-result-dnn1}
  \centering
  \scalebox{1}{
  \begin{tabular}{cccccccc}
    \toprule
    \multirow{1}{*}{Algorithm} & $\eta_1$ & $\eta_2$ & $\lambda$ & $\alpha$ &  PM Acc.(\%) &  GM Acc.(\%) & Personalized\\
    \midrule
        \multirow{3}{*}
        pFedMe & 0.0005 & 0.001 & 15 & -& 83.22 & 80.86 & Y\\
        pFedMe & 0.001 & 0.001 & 15 & - & 89.35 & 86.86 & Y\\
        pFedMe & 0.005 & 0.005 & 15 & - & 94.35 & 89.17 & Y\\
        \textbf{pFedMe} & \textbf{0.01}  & \textbf{0.01}  & \textbf{15} & - & \textbf{95.16} & \textbf{89.31} & Y\\
        pFedMe & 0.1 & 0.1 & 15 & - & 19.73 & 9.9 & Y\\
    \midrule
        \multirow{3}{*}
        HeurFedAMP & 0.0005 & - & - & 0.1 & 89.73 & - & Y\\
        HeurFedAMP & 0.001 & - & - & 0.1 & 89.91 & - & Y\\
        HeurFedAMP & 0.005 & - & - & 0.1 & 89.41 & - & Y\\
        HeurFedAMP & 0.01 & - & - & 0.1 & 88.68 & - & Y\\
        HeurFedAMP & 0.1 & - & - & 0.1 & 10.03 & - & Y\\
    \midrule
        \multirow{3}{*}
        HeurFedAMP & 0.0005 & - & - & 0.5 & 93.32 & - & Y\\
        HeurFedAMP & 0.001 & - & - & 0.5 & 94.04 & - & Y\\
        HeurFedAMP & 0.005 & - & - & 0.5 & 94.90 & - & Y\\
        \textbf{HeurFedAMP} & \textbf{0.01} & - & - & \textbf{0.5} & \textbf{94.95} & - & Y\\
        HeurFedAMP & 0.1 & - & - & 0.5 & 93.29 & - & Y\\
    \midrule
        \multirow{3}{*}
        Per-FedAvg & 0.0005 & - & - & - & 94.71 & - & Y\\
        Per-FedAvg & 0.001 & - & - & - & 94.95 & - & Y\\
        Per-FedAvg & 0.005 & - & - & - & 95.31 & - & Y\\
        \textbf{Per-FedAvg} & \textbf{0.01} & - & \textbf{-} & - & \textbf{95.35} & \textbf{-} & Y\\
        Per-FedAvg & 0.1 & - & - & - & 19.94 & - & Y\\
    
    \midrule
        \multirow{3}{*}
        pFedGP & 0.01 & - & - & - & 90.52 & - & Y\\
        \textbf{pFedGP} & \textbf{0.05} & - & - & \textbf{-} & \textbf{92.14} & - & Y\\
        pFedGP & 0.1 & - & - & - & 91.23 & - & Y\\ 
    \midrule
        \multirow{3}{*}
        FedAvg & - & 0.0005 & - & - & - & 87.65 & N\\
        FedAvg & - & 0.001 & - & - & - & 89.19 & N\\
        FedAvg & - & 0.005 & - & - & - & 90.08 & N\\
        \textbf{FedAvg} & - & \textbf{0.01} & \textbf{-} & \textbf{-} & \textbf{-} & \textbf{90.66} & N\\
    \midrule
        \multirow{3}{*}
        Fedprox & - & 0.001 & 0.001 & - & - & 89.26 & N\\
        Fedprox & - & 0.001 & 0.01 & - & - & 89.24 & N\\
        Fedprox  & - & 0.001  & 0.1 & - & - & 87.79 & N\\
        Fedprox  & - & 0.001  & 1 & - & - & 76.79 & N\\
    \midrule
        \multirow{3}{*}
        Fedprox & - & 0.0005 & 0.001 & - & - & 87.64 & N\\
        Fedprox & - & 0.001 & 0.001 & - & - & 89.26 & N\\
        Fedprox & - & 0.005 & 0.001 & - & - & 89.84 & N\\
        \textbf{Fedprox} & - & \textbf{0.01} & \textbf{0.001} & \textbf{-} & \textbf{-} & \textbf{90.70} & N\\
    \midrule
        \multirow{3}{*}
        BNFed & - & 0.0005 & - & - & - & 9.90 & N\\
        BNFed & - & 0.001 & - & - & - & 9.90 & N\\
        BNFed & - & 0.005 & - & - & - & 9.90 & N\\
        BNFed & - & 0.01 & - & - & - & 9.94 & N\\
        BNFed & - & 0.1 & - & - & - & 55.98 & N\\
        BNFed & - & 0.2 & - & - & - & 74.47 & N\\
        \textbf{BNFed} & - & \textbf{0.5} & \textbf{-} & - & - & \textbf{80.31} & N\\
    \bottomrule
  \end{tabular}}
\end{table*}

\subsection{{\texttt{pFedBayes}} Uncertainty Estimation}

One of the advantages of Bayesian algorithm is that it can provide uncertainty estimation results. Figure~\ref{fig:fig2} shows that the predication uncertainty of personalized model of our {\texttt{pFedBayes}}. We can see that random model parameters do not give accurate estimates before training. As training progresses, the model becomes more and more confident in the classification results. The results of uncertainty estimation can give people a good reference, especially in federated learning architectures. When aggregating models, we can know the uncertainty of each client model. This information can be used to evaluate the quality of the client model, decide whether to aggregate, etc.


\begin{thebibliography}{55}
	\providecommand{\natexlab}[1]{#1}
	\providecommand{\url}[1]{\texttt{#1}}
	\expandafter\ifx\csname urlstyle\endcsname\relax
	\providecommand{\doi}[1]{doi: #1}\else
	\providecommand{\doi}{doi: \begingroup \urlstyle{rm}\Url}\fi
	
	\bibitem[Achituve et~al.(2021)Achituve, Shamsian, Navon, Chechik, and
	Fetaya]{achituve2021personalized}
	Achituve, I., Shamsian, A., Navon, A., Chechik, G., and Fetaya, E.
	\newblock Personalized federated learning with gaussian processes.
	\newblock In \emph{Thirty-Fifth Conference on Neural Information Processing
		Systems}, 2021.
	
	\bibitem[Al-Shedivat et~al.(2021)Al-Shedivat, Gillenwater, Xing, and
	Rostamizadeh]{al2021federated}
	Al-Shedivat, M., Gillenwater, J., Xing, E., and Rostamizadeh, A.
	\newblock Federated learning via posterior averaging: A new perspective and
	practical algorithms.
	\newblock In \emph{International Conference on Learning Representations}, 2021.
	
	\bibitem[Alquier \& Ridgway(2020)Alquier and Ridgway]{alquier2020concentration}
	Alquier, P. and Ridgway, J.
	\newblock Concentration of tempered posteriors and of their variational
	approximations.
	\newblock \emph{The Annals of Statistics}, 48\penalty0 (3):\penalty0
	1475--1497, 2020.
	
	\bibitem[Arivazhagan et~al.(2019)Arivazhagan, Aggarwal, Singh, and
	Choudhary]{arivazhagan2019federated}
	Arivazhagan, M.~G., Aggarwal, V., Singh, A.~K., and Choudhary, S.
	\newblock Federated learning with personalization layers.
	\newblock \emph{arXiv preprint arXiv:1912.00818}, 2019.
	
	\bibitem[Bai et~al.(2020)Bai, Song, and Cheng]{bai2020efficient}
	Bai, J., Song, Q., and Cheng, G.
	\newblock Efficient variational inference for sparse deep learning with
	theoretical guarantee.
	\newblock \emph{Advances in Neural Information Processing Systems}, 33, 2020.
	
	\bibitem[Blei et~al.(2017)Blei, Kucukelbir, and McAuliffe]{Blei2017ReviewVB}
	Blei, D.~M., Kucukelbir, A., and McAuliffe, J.~D.
	\newblock Variational inference: A review for statisticians.
	\newblock \emph{Journal of the American Statistical Association}, 112\penalty0
	(518):\penalty0 859--877, 2017.
	
	\bibitem[Blundell et~al.(2015{\natexlab{a}})Blundell, Cornebise, Kavukcuoglu,
	and Wierstra]{Blundell2015}
	Blundell, C., Cornebise, J., Kavukcuoglu, K., and Wierstra, D.
	\newblock Weight uncertainty in neural networks.
	\newblock In \emph{Proceedings of the 32Nd International Conference on
		International Conference on Machine Learning - Volume 37}, ICML'15, pp.\
	1613--1622. JMLR.org, 2015{\natexlab{a}}.
	
	\bibitem[Blundell et~al.(2015{\natexlab{b}})Blundell, Cornebise, Kavukcuoglu,
	and Wierstra]{blundell2015weight}
	Blundell, C., Cornebise, J., Kavukcuoglu, K., and Wierstra, D.
	\newblock Weight uncertainty in neural network.
	\newblock In \emph{International Conference on Machine Learning}, pp.\
	1613--1622. PMLR, 2015{\natexlab{b}}.
	
	\bibitem[Boucheron et~al.(2013)Boucheron, Lugosi, and
	Massart]{Boucheron2013Concentration}
	Boucheron, S., Lugosi, G., and Massart, P.
	\newblock \emph{Concentration inequalities: A nonasymptotic theory of
		independence}.
	\newblock Oxford university press, 2013.
	
	\bibitem[Chen et~al.(2018)Chen, Luo, Dong, Li, and He]{chen2018federated}
	Chen, F., Luo, M., Dong, Z., Li, Z., and He, X.
	\newblock Federated meta-learning with fast convergence and efficient
	communication.
	\newblock \emph{arXiv preprint arXiv:1802.07876}, 2018.
	
	\bibitem[Chen \& Chao(2021)Chen and Chao]{chen2020fedbe}
	Chen, H.-Y. and Chao, W.-L.
	\newblock {FedBE: Making Bayesian} model ensemble applicable to federated
	learning.
	\newblock In \emph{International Conference on Learning Representations}, 2021.
	
	\bibitem[Ch{\'e}rief-Abdellatif(2020)]{cherief2020convergence}
	Ch{\'e}rief-Abdellatif, B.-E.
	\newblock Convergence rates of variational inference in sparse deep learning.
	\newblock In \emph{International Conference on Machine Learning}, pp.\
	1831--1842. PMLR, 2020.
	
	\bibitem[Ch{\'e}rief-Abdellatif \& Alquier(2018)Ch{\'e}rief-Abdellatif and
	Alquier]{cherief2018consistency}
	Ch{\'e}rief-Abdellatif, B.-E. and Alquier, P.
	\newblock Consistency of variational bayes inference for estimation and model
	selection in mixtures.
	\newblock \emph{Electronic Journal of Statistics}, 12\penalty0 (2):\penalty0
	2995--3035, 2018.
	
	\bibitem[Dai et~al.(2019)Dai, Yan, Zhou, Yang, Ng, Cheng, and
	Fan]{dai2019hyper}
	Dai, X., Yan, X., Zhou, K., Yang, H., Ng, K.~K., Cheng, J., and Fan, Y.
	\newblock Hyper-sphere quantization: Communication-efficient sgd for federated
	learning.
	\newblock \emph{arXiv preprint arXiv:1911.04655}, 2019.
	
	\bibitem[Fallah et~al.(2020)Fallah, Mokhtari, and
	Ozdaglar]{fallah2020personalized}
	Fallah, A., Mokhtari, A., and Ozdaglar, A.
	\newblock Personalized federated learning with theoretical guarantees: A
	model-agnostic meta-learning approach.
	\newblock \emph{Advances in Neural Information Processing Systems},
	33:\penalty0 3557--3568, 2020.
	
	\bibitem[Guha et~al.(2019)Guha, Talwalkar, and Smith]{guha2019one}
	Guha, N., Talwalkar, A., and Smith, V.
	\newblock One-shot federated learning.
	\newblock \emph{arXiv preprint arXiv:1902.11175}, 2019.
	
	\bibitem[Hanzely \& Richt{\'a}rik(2020)Hanzely and
	Richt{\'a}rik]{hanzely2020federated}
	Hanzely, F. and Richt{\'a}rik, P.
	\newblock Federated learning of a mixture of global and local models.
	\newblock \emph{arXiv preprint arXiv:2002.05516}, 2020.
	
	\bibitem[Higgins et~al.(2017)Higgins, Matthey, Pal, Burgess, Glorot, Botvinick,
	Mohamed, and Lerchner]{higgins2017beta}
	Higgins, I., Matthey, L., Pal, A., Burgess, C., Glorot, X., Botvinick, M.,
	Mohamed, S., and Lerchner, A.
	\newblock Beta-{VAE}: Learning basic visual concepts with a constrained
	variational framework.
	\newblock In \emph{International Conference on Learning Representations}, 2017.
	
	\bibitem[Huang et~al.(2021)Huang, Chu, Zhou, Wang, Liu, Pei, and
	Zhang]{huang2021personalized}
	Huang, Y., Chu, L., Zhou, Z., Wang, L., Liu, J., Pei, J., and Zhang, Y.
	\newblock Personalized cross-silo federated learning on non-iid data.
	\newblock In \emph{Proceedings of the AAAI Conference on Artificial
		Intelligence}, volume~35, pp.\  7865--7873, 2021.
	
	\bibitem[Jordan et~al.(1999)Jordan, Ghahramani, Jaakkola, and
	Saul]{VIJordan1999}
	Jordan, M.~I., Ghahramani, Z., Jaakkola, T.~S., and Saul, L.~K.
	\newblock An introduction to variational methods for graphical models.
	\newblock \emph{Machine Learning}, 37:\penalty0 183--233, 1999.
	
	\bibitem[Jospin et~al.(2020)Jospin, Buntine, Boussaid, Laga, and
	Bennamoun]{jospin2020hands}
	Jospin, L.~V., Buntine, W., Boussaid, F., Laga, H., and Bennamoun, M.
	\newblock Hands-on bayesian neural networks-a tutorial for deep learning users.
	\newblock \emph{ACM Comput. Surv}, 1\penalty0 (1), 2020.
	
	\bibitem[Karimireddy et~al.(2020)Karimireddy, Kale, Mohri, Reddi, Stich, and
	Suresh]{karimireddy2020scaffold}
	Karimireddy, S.~P., Kale, S., Mohri, M., Reddi, S., Stich, S., and Suresh,
	A.~T.
	\newblock Scaffold: Stochastic controlled averaging for federated learning.
	\newblock In \emph{International Conference on Machine Learning}, pp.\
	5132--5143. PMLR, 2020.
	
	\bibitem[Khan(2019)]{khan2019deep}
	Khan, M.~E.
	\newblock Deep learning with bayesian principles.
	\newblock \emph{Tutorial on Advances in Neural Information Processing Systems},
	2019.
	
	\bibitem[Krizhevsky(2009)]{krizhevsky2009learning}
	Krizhevsky, A.
	\newblock Learning multiple layers of features from tiny images.
	\newblock \emph{Master's thesis, University of Tront}, 2009.
	
	\bibitem[LeCun et~al.(1998)LeCun, Bottou, Bengio, and
	Haffner]{lecun1998gradient}
	LeCun, Y., Bottou, L., Bengio, Y., and Haffner, P.
	\newblock Gradient-based learning applied to document recognition.
	\newblock \emph{Proceedings of the IEEE}, 86\penalty0 (11):\penalty0
	2278--2324, 1998.
	
	\bibitem[LeCun et~al.(2010)LeCun, Cortes, and Burges]{lecun2010mnist}
	LeCun, Y., Cortes, C., and Burges, C.
	\newblock Mnist handwritten digit database.
	\newblock \emph{ATT Labs [Online]. Available:
		http://yann.lecun.com/exdb/mnist}, 2, 2010.
	
	\bibitem[Li et~al.(2018)Li, Sahu, Zaheer, Sanjabi, Talwalkar, and
	Smith]{li2018federated}
	Li, T., Sahu, A.~K., Zaheer, M., Sanjabi, M., Talwalkar, A., and Smith, V.
	\newblock Federated optimization in heterogeneous networks.
	\newblock \emph{arXiv preprint arXiv:1812.06127}, 2018.
	
	\bibitem[Li et~al.(2019)Li, Sahu, Zaheer, Sanjabi, Talwalkar, and
	Smithy]{li2019feddane}
	Li, T., Sahu, A.~K., Zaheer, M., Sanjabi, M., Talwalkar, A., and Smithy, V.
	\newblock {Feddane: A} federated newton-type method.
	\newblock In \emph{2019 53rd Asilomar Conference on Signals, Systems, and
		Computers}, pp.\  1227--1231. IEEE, 2019.
	
	\bibitem[Li et~al.(2020)Li, Sahu, Talwalkar, and Smith]{Li2020FL}
	Li, T., Sahu, A.~K., Talwalkar, A., and Smith, V.
	\newblock Federated learning: Challenges, methods, and future directions.
	\newblock \emph{IEEE Signal Processing Magazine}, 37\penalty0 (3):\penalty0
	50--60, 2020.
	
	\bibitem[Li et~al.(2022{\natexlab{a}})Li, Xu, Song, Li, Li, Shao, and
	Zhan]{li2022federated}
	Li, X.-C., Xu, Y.-C., Song, S., Li, B., Li, Y., Shao, Y., and Zhan, D.-C.
	\newblock Federated learning with position-aware neurons.
	\newblock In \emph{Proceedings of the IEEE/CVF Conference on Computer Vision
		and Pattern Recognition}, pp.\  10082--10091, 2022{\natexlab{a}}.
	
	\bibitem[Li et~al.(2021)Li, Liu, Zhang, Shao, Wang, and
	Geng]{li2021personalized}
	Li, Y., Liu, X., Zhang, X., Shao, Y., Wang, Q., and Geng, Y.
	\newblock Personalized federated learning via maximizing correlation with
	sparse and hierarchical extensions.
	\newblock \emph{arXiv preprint arXiv:2107.05330}, 2021.
	
	\bibitem[Li et~al.(2022{\natexlab{b}})Li, Lu, Luo, Zhu, Shao, Li, Zhang, and
	Wu]{li2022mining}
	Li, Z., Lu, J., Luo, S., Zhu, D., Shao, Y., Li, Y., Zhang, Z., and Wu, C.
	\newblock Mining latent relationships among clients: Peer-to-peer federated
	learning with adaptive neighbor matching.
	\newblock \emph{arXiv preprint arXiv:2203.12285}, 2022{\natexlab{b}}.
	
	\bibitem[Liu et~al.(2021)Liu, Zheng, Chen, Qi, Huang, and
	Shao]{liu2021bayesian}
	Liu, L., Zheng, F., Chen, H., Qi, G.-J., Huang, H., and Shao, L.
	\newblock A bayesian federated learning framework with online laplace
	approximation.
	\newblock \emph{arXiv preprint arXiv:2102.01936}, 2021.
	
	\bibitem[Liu et~al.(2022)Liu, Li, Shao, and Wang]{liu2022sparse}
	Liu, X., Li, Y., Shao, Y., and Wang, Q.
	\newblock Sparse federated learning with hierarchical personalization models.
	\newblock \emph{arXiv preprint arXiv:2203.13517}, 2022.
	
	\bibitem[MacKay(1992)]{mackay1992practical}
	MacKay, D.~J.
	\newblock A practical bayesian framework for backpropagation networks.
	\newblock \emph{Neural computation}, 4\penalty0 (3):\penalty0 448--472, 1992.
	
	\bibitem[Maddox et~al.(2019)Maddox, Izmailov, Garipov, Vetrov, and
	Wilson]{maddox2019simple}
	Maddox, W.~J., Izmailov, P., Garipov, T., Vetrov, D.~P., and Wilson, A.~G.
	\newblock A simple baseline for bayesian uncertainty in deep learning.
	\newblock \emph{Advances in Neural Information Processing Systems},
	32:\penalty0 13153--13164, 2019.
	
	\bibitem[McMahan et~al.(2017)McMahan, Moore, Ramage, Hampson, and
	y~Arcas]{mcmahan2017communication}
	McMahan, B., Moore, E., Ramage, D., Hampson, S., and y~Arcas, B.~A.
	\newblock Communication-efficient learning of deep networks from decentralized
	data.
	\newblock In \emph{Artificial Intelligence and Statistics}, pp.\  1273--1282.
	PMLR, 2017.
	
	\bibitem[Nakada \& Imaizumi(2020)Nakada and Imaizumi]{nakada2020}
	Nakada, R. and Imaizumi, M.
	\newblock Adaptive approximation and generalization of deep neural network with
	intrinsic dimensionality.
	\newblock \emph{Journal of Machine Learning Research}, 21\penalty0
	(174):\penalty0 1--38, 2020.
	
	\bibitem[Neal(2012)]{neal2012bayesian}
	Neal, R.~M.
	\newblock \emph{Bayesian learning for neural networks}, volume 118.
	\newblock Springer Science \& Business Media, 2012.
	
	\bibitem[Osawa et~al.(2019)Osawa, Swaroop, Khan, Jain, Eschenhagen, Turner, and
	Yokota]{Osawa2019Practical}
	Osawa, K., Swaroop, S., Khan, M. E.~E., Jain, A., Eschenhagen, R., Turner,
	R.~E., and Yokota, R.
	\newblock Practical deep learning with bayesian principles.
	\newblock \emph{Advances in neural information processing systems}, 32, 2019.
	
	\bibitem[Paszke et~al.(2019)Paszke, Gross, Massa, Lerer, Bradbury, Chanan,
	Killeen, Lin, Gimelshein, Antiga, Desmaison, Kopf, Yang, DeVito, Raison,
	Tejani, Chilamkurthy, Steiner, Fang, Bai, and Chintala]{NEURIPS2019_bdbca288}
	Paszke, A., Gross, S., Massa, F., Lerer, A., Bradbury, J., Chanan, G., Killeen,
	T., Lin, Z., Gimelshein, N., Antiga, L., Desmaison, A., Kopf, A., Yang, E.,
	DeVito, Z., Raison, M., Tejani, A., Chilamkurthy, S., Steiner, B., Fang, L.,
	Bai, J., and Chintala, S.
	\newblock Pytorch: An imperative style, high-performance deep learning library.
	\newblock In Wallach, H., Larochelle, H., Beygelzimer, A., d\textquotesingle
	Alch\'{e}-Buc, F., Fox, E., and Garnett, R. (eds.), \emph{Advances in Neural
		Information Processing Systems}, volume~32. Curran Associates, Inc., 2019.
	
	\bibitem[Pati et~al.(2018)Pati, Bhattacharya, and Yang]{pati2018statistical}
	Pati, D., Bhattacharya, A., and Yang, Y.
	\newblock On statistical optimality of variational bayes.
	\newblock In \emph{International Conference on Artificial Intelligence and
		Statistics}, pp.\  1579--1588. PMLR, 2018.
	
	\bibitem[Polson \& Ro{\v{c}}kov{\'a}(2018)Polson and
	Ro{\v{c}}kov{\'a}]{polson2018posterior}
	Polson, N.~G. and Ro{\v{c}}kov{\'a}, V.
	\newblock Posterior concentration for sparse deep learning.
	\newblock In \emph{Proceedings of the 32nd International Conference on Neural
		Information Processing Systems}, pp.\  938--949, 2018.
	
	\bibitem[Reisizadeh et~al.(2020)Reisizadeh, Mokhtari, Hassani, Jadbabaie, and
	Pedarsani]{reisizadeh2020fedpaq}
	Reisizadeh, A., Mokhtari, A., Hassani, H., Jadbabaie, A., and Pedarsani, R.
	\newblock Fedpaq: A communication-efficient federated learning method with
	periodic averaging and quantization.
	\newblock In \emph{International Conference on Artificial Intelligence and
		Statistics}, pp.\  2021--2031. PMLR, 2020.
	
	\bibitem[Rothchild et~al.(2020)Rothchild, Panda, Ullah, Ivkin, Stoica,
	Braverman, Gonzalez, and Arora]{rothchild2020fetchsgd}
	Rothchild, D., Panda, A., Ullah, E., Ivkin, N., Stoica, I., Braverman, V.,
	Gonzalez, J., and Arora, R.
	\newblock {Fetchsgd: C}ommunication-efficient federated learning with
	sketching.
	\newblock In \emph{International Conference on Machine Learning}, pp.\
	8253--8265. PMLR, 2020.
	
	\bibitem[Sattler et~al.(2019)Sattler, Wiedemann, M{\"u}ller, and
	Samek]{sattler2019robust}
	Sattler, F., Wiedemann, S., M{\"u}ller, K.-R., and Samek, W.
	\newblock Robust and communication-efficient federated learning from non-iid
	data.
	\newblock \emph{IEEE transactions on neural networks and learning systems},
	31\penalty0 (9):\penalty0 3400--3413, 2019.
	
	\bibitem[Sattler et~al.(2021)Sattler, Müller, and Samek]{sattler2021clustered}
	Sattler, F., Müller, K.-R., and Samek, W.
	\newblock Clustered federated learning: Model-agnostic distributed multitask
	optimization under privacy constraints.
	\newblock \emph{IEEE Transactions on Neural Networks and Learning Systems},
	32\penalty0 (8):\penalty0 3710--3722, 2021.
	\newblock \doi{10.1109/TNNLS.2020.3015958}.
	
	\bibitem[Simonyan \& Zisserman(2014)Simonyan and Zisserman]{simonyan2014very}
	Simonyan, K. and Zisserman, A.
	\newblock Very deep convolutional networks for large-scale image recognition.
	\newblock \emph{arXiv preprint arXiv:1409.1556}, 2014.
	
	\bibitem[Smith et~al.(2017)Smith, Chiang, Sanjabi, and
	Talwalkar]{smith2017federated}
	Smith, V., Chiang, C.-K., Sanjabi, M., and Talwalkar, A.~S.
	\newblock Federated multi-task learning.
	\newblock In \emph{Advances in neural information processing systems}, pp.\
	4424--4434, 2017.
	
	\bibitem[T~Dinh et~al.(2020)T~Dinh, Tran, and Nguyen]{t2020personalized}
	T~Dinh, C., Tran, N., and Nguyen, T.~D.
	\newblock Personalized federated learning with moreau envelopes.
	\newblock \emph{Advances in Neural Information Processing Systems}, 33, 2020.
	
	\bibitem[Thorgeirsson \& Gauterin(2021)Thorgeirsson and
	Gauterin]{thorgeirsson2021probabilistic}
	Thorgeirsson, A.~T. and Gauterin, F.
	\newblock Probabilistic predictions with federated learning.
	\newblock \emph{Entropy}, 23\penalty0 (1):\penalty0 41, 2021.
	
	\bibitem[Xiao et~al.(2017)Xiao, Rasul, and Vollgraf]{xiao2017fashion}
	Xiao, H., Rasul, K., and Vollgraf, R.
	\newblock Fashion-mnist: a novel image dataset for benchmarking machine
	learning algorithms.
	\newblock \emph{arXiv preprint arXiv:1708.07747}, 2017.
	
	\bibitem[Yurochkin et~al.(2019)Yurochkin, Agarwal, Ghosh, Greenewald, Hoang,
	and Khazaeni]{yurochkin2019bayesian}
	Yurochkin, M., Agarwal, M., Ghosh, S., Greenewald, K., Hoang, N., and Khazaeni,
	Y.
	\newblock Bayesian nonparametric federated learning of neural networks.
	\newblock In \emph{International Conference on Machine Learning}, pp.\
	7252--7261. PMLR, 2019.
	
	\bibitem[Zhang et~al.(2020)Zhang, Hong, Dhople, Yin, and Liu]{zhang2020fedpd}
	Zhang, X., Hong, M., Dhople, S., Yin, W., and Liu, Y.
	\newblock {Fedpd: A} federated learning framework with optimal rates and
	adaptivity to non-iid data.
	\newblock \emph{arXiv preprint arXiv:2005.11418}, 2020.
	
	\bibitem[Zong et~al.(2021)Zong, Wang, Liu, Li, and Shao]{zong2021communication}
	Zong, H., Wang, Q., Liu, X., Li, Y., and Shao, Y.
	\newblock Communication reducing quantization for federated learning with local
	differential privacy mechanism.
	\newblock In \emph{2021 IEEE/CIC International Conference on Communications in
		China (ICCC)}, pp.\  75--80. IEEE, 2021.
	
\end{thebibliography}
\end{document}